\pdfoutput=1

\documentclass[11pt]{article}

\usepackage[]{acl}
\usepackage{times}
\usepackage{latexsym}
\usepackage[T1]{fontenc}
\usepackage[utf8]{inputenc}
\usepackage{microtype}
\usepackage{booktabs, tabularx, multirow, graphics, boldline, hhline}
\usepackage{array, amssymb, amsmath, enumitem}
\usepackage{graphicx, color, soul, xcolor}
\usepackage{cite}
\usepackage{xr}
\usepackage{hyperref}
\usepackage[normalem]{ulem}
\usepackage{pifont}% http://ctan.org/pkg/pifont
\newcommand{\xmark}{\ding{55}}%
\title{On the Use of External Data for Spoken Named Entity Recognition}
\author{
Ankita Pasad\thanks{\;\;Work done during an internship at ASAPP.}\;\;$^{2}$, Felix Wu$^{1}$, Suwon Shon$^{1}$, Karen Livescu$^{2}$, Kyu J. Han$^{1}$ \\
 $^1$ASAPP \ \ \ \ \   $^2$Toyota Technological Institute at Chicago \\
}
\begin{document}
\maketitle
\begin{abstract}
Spoken language understanding (SLU) tasks involve mapping from speech signals to semantic labels. Given the complexity of such tasks, good performance is expected to require large labeled datasets, which are difficult to collect for each new task and domain. However, recent advances in self-supervised speech representations have made it feasible to consider learning SLU models with limited labeled data. In this work, we focus on low-resource spoken named entity recognition (NER) and address the question: Beyond self-supervised pre-training, how can we use external speech and/or text data that are not annotated for the task? We consider self-training, knowledge distillation, and transfer learning for end-to-end (E2E) and pipeline (speech recognition followed by text NER) approaches. We find that several of these approaches improve performance in resource-constrained settings beyond the benefits from pre-trained representations. Compared to prior work, we find relative improvements in F1 of up to 16\%. While the best baseline model is a pipeline approach, the best performance using external data is ultimately achieved by an E2E model. We provide detailed comparisons and analyses, developing insights on, for example, the effects of leveraging external data on (i) different categories of NER errors and (ii) the switch in performance trends between pipeline and E2E models. Code is available at \href{https://github.com/asappresearch/spoken-ner}{https://github.com/asappresearch/spoken-ner}.

\end{abstract}
% \ap{Limit: 8 pages for the main content (no limit on references)}

\section{Introduction}
\label{sec:intro}
Named entity recognition (NER) is a popular task in natural language processing. It involves detecting the named entities and their categories from a text sequence. NER can be used to extract information from unstructured data, which can also be used as features for other tasks like question answering~\citep{chen2017reading} and slot filling for task-oriented dialogues~\citep{louvan-magnini-2018-exploring}. 

\begin{figure}[t]
    \centering
    \hspace{-.15in}
    \includegraphics[width=8cm, trim=0 65 0 80]{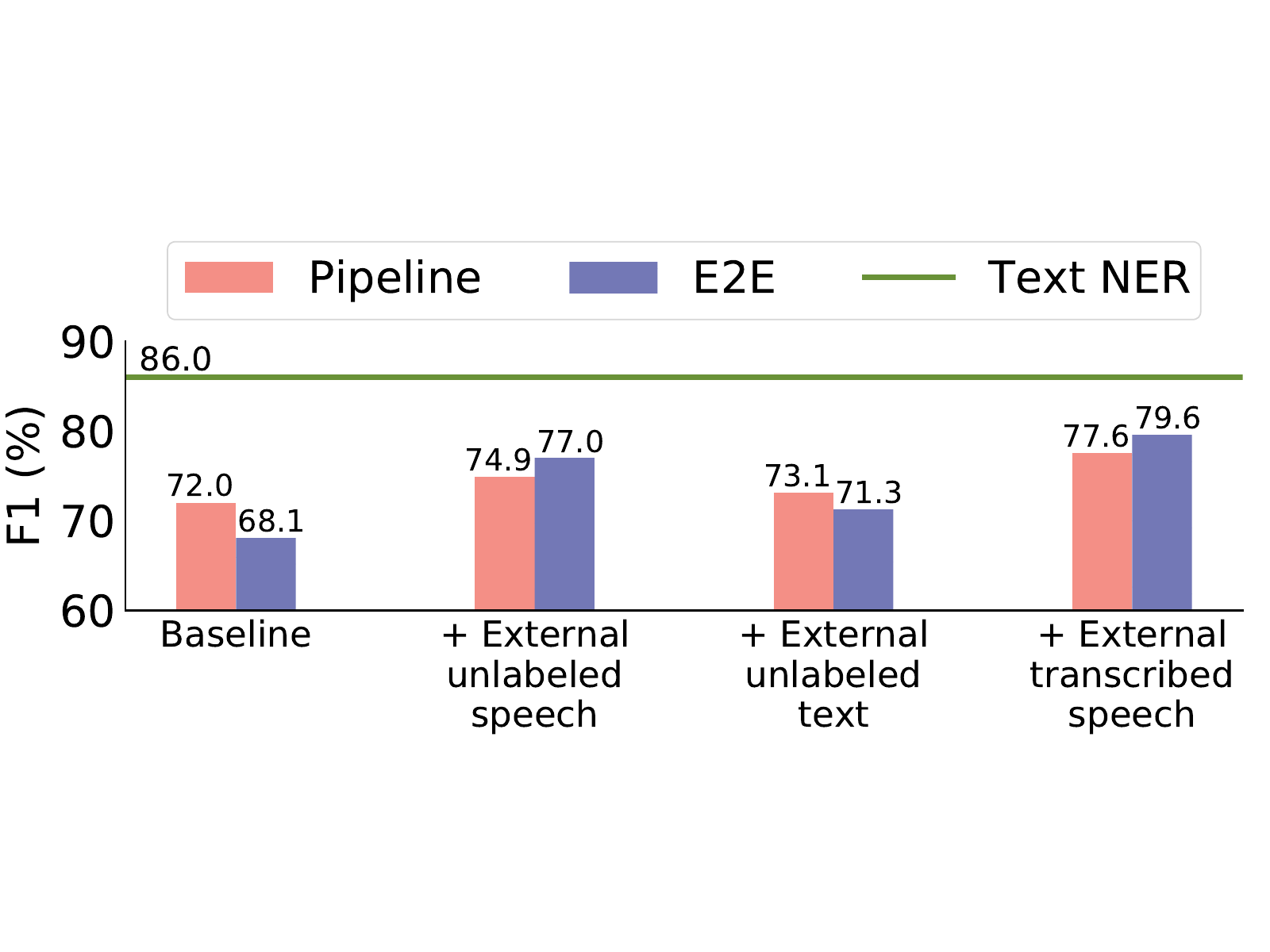}
    \caption{\it Improvements in spoken NER with 100 hours of external data of different types. ``Pipeline'' refers to approaches consisting of speech recognition followed by a text NER model; ``E2E'' refers to approaches that directly map from speech to NER-tagged text. The ``Baseline'' and ``Text NER'' numbers are from previously established baselines~\citep{slue}.}
    \label{fig:summary}
\end{figure}

Thanks to pre-trained text representations such as BERT~\citep{devlin2018bert}, text-based NER has recently improved greatly~\citep{wang2020automated, li2019dice}. {\it Spoken} NER, on the other hand, is not as well-studied. It has the added challenges of continuous-valued and longer input sequences and, at the same time, provides opportunities to take advantage of acoustic cues in the input. A recent study~\citep{slue} shows that there is still 10-20\% absolute degradation in the F1 scores of spoken NER models compared to text-based NER using gold transcripts (see \autoref{fig:summary}) despite using large pre-trained speech representation models. Closing this gap remains a critical challenge.

In this work, we study the potential benefits of using a variety of external data types: (a) plain speech audio, (b) plain text, (c) speech with transcripts, and (d) text-based NER data. We benchmark our findings against recently published baselines for NER on the VoxPopuli dataset of European Parliament speech recordings~\citep{slue} and also introduce baselines of our own. We observe improvement from leveraging every type of external data. Our analysis also quantifies the pros and cons of the pipeline (speech recognition followed by text NER) and end-to-end (E2E) approaches. The key improvements are summarized in Figure~\ref{fig:summary}. Specific contributions include: \\
(i) Unlike previous work, we devote equal effort to improving both pipeline and E2E approaches. \\
(ii) We present experiments using various external data types and modeling approaches. \\
(iii) Overall, we obtain F1 improvements of up to 16\% for the E2E model and 6\% for the pipeline model over previously published baselines, setting a new state of the art for NER on this dataset. \\
(iv) We benchmark the advantage of self-supervised representations (SSR) over a baseline that uses standard spectrogram features. SSR gives relative improvements of 36\%/31\% for pipeline/E2E models, respectively. To our knowledge, prior work has not directly measured this improvement over competitive baselines tuned for the task. \\
(v) We establish that E2E models outperform pipeline approaches on this task, given access to external data, while the baseline models without the external data have the opposite relationship. \\
(vi) We provide a detailed analysis of model behavior, including differences in error types between pipeline and E2E approaches and the reasoning for the superiority of E2E over pipeline models when using external data but not in the baseline setting.

\section{Related work} \label{sec:related_work}
\begin{figure*}[ht]
  \centering
  \includegraphics[width=\linewidth]{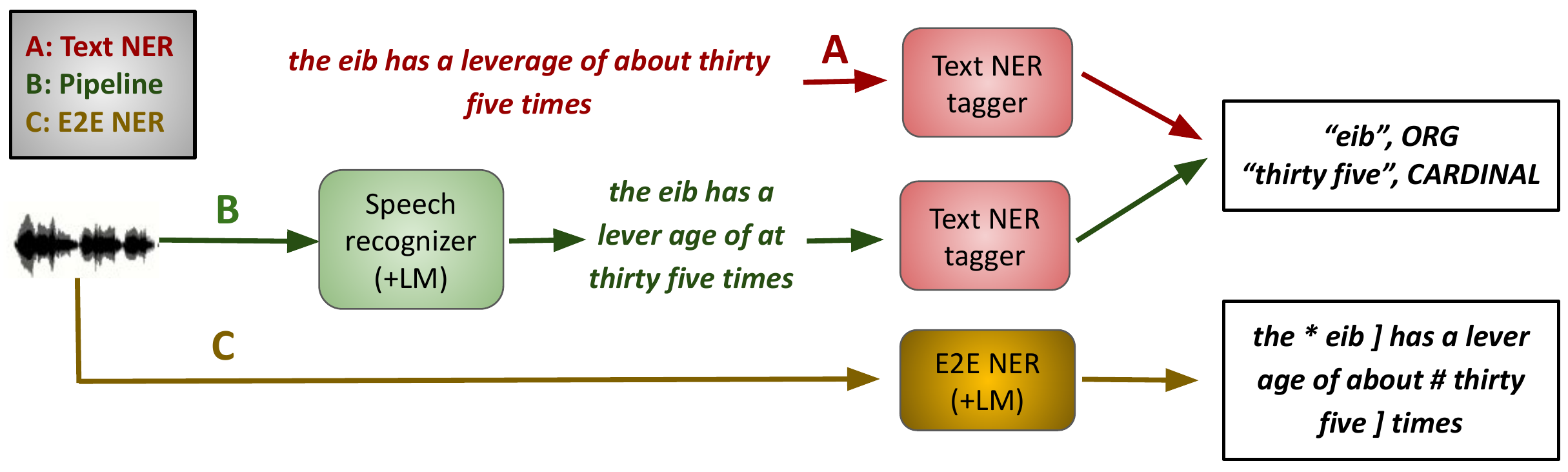}
  \caption{\it High-level summary of approaches typically used to solve spoken and textual NER tasks. Optional LM decoding is applied in ASR and E2E-NER models.} 
  \label{fig:baseline-systems}
\end{figure*}

\subsection{Spoken named entity recognition}
\label{sec:spoken-ner}
Relatively little work has been conducted on spoken NER~\citep{kim2000rule,sudoh2006incorporating,parada2011oov,ghannay2018end,caubriere2020we,yadav2020end,slue} as compared to the extensively studied task of NER on text~\citep{nadeau2007survey,ratinov2009design,yadav2019survey, li2020survey}.
% ~\citep{mikheev1999named,florian2003named,nadeau2007survey,ratinov2009design,ritter2011named,lample2016neural,chiu2016named,akbik2019naacl,wang2020automated,yamada2020luke}
While spoken NER is commonly done through a pipeline approach~\citep{sudoh2006incorporating,raymond2013robust,jannet2015evaluate}, there is rising interest in E2E approaches in the speech community
~\citep{ghannay2018end,caubriere2020we,yadav2020end,slue}. These two approaches are depicted in Fig.~\ref{fig:baseline-systems}.

An early E2E spoken NER model was introduced by~\citet{ghannay2018end}. The approach is based on the DeepSpeech2~\citep{amodei2016deep} architecture, with the addition of special characters for NER labels around the named entities in the transcription, and is trained with character-level connectionist temporal classification (CTC)~\citep{graves2006connectionist}. \citet{yadav2020end} introduced an English speech NER dataset and proposed an E2E approach similar to \citet{ghannay2018end}. They show that LM fusion improves the performance of the E2E approach. \citet{caubriere2020we} provided a detailed comparison between E2E and pipeline models; however, they focused on small RNN/CNN models and did not use state-of-the-art SSR models. All these approaches use at least 100 hours of annotated data.

These previous efforts have shown that E2E models can outperform pipeline approaches in a fully supervised setting. \citet{borgholt2021we} also made the same observation on a simplified NER task. However, these studies do not account for improvements in NLP from self-supervised text representations for their pipeline counterparts. \citet{slue} introduced and worked with a low-resource NER corpus and showed that E2E models still do not rival pipeline approaches when state-of-the-art pre-trained models are used.

When using pre-trained representations, E2E models are at a disadvantage since the pipeline model also has access to a text model trained on $>$50GB of text, in addition to the same speech representation model as E2E. This inspires us to study the benefits of using additional unlabeled data. 

We choose to work with the NER-annotated VoxPopuli corpus~\citep{wang2021voxpopuli, slue}. VoxPopuli consists of naturally spoken speech, unlike \citet{bastianelli2020slurp}, and is annotated manually, unlike \citet{yadav2020end} and \citet{borgholt2021we} who obtain ground-truth labels using text model predictions. The SLUE benchmark~\citep{slue} is aimed at low-resource SLU and includes annotations for only 15 hours of data; this matches with the goals of our work, making it an ideal choice to benchmark our findings.

\subsection{Self-supervised pre-training}
There is a long history of using unsupervised pre-training in NLP to improve performance over limited-data supervised training on a broad range of tasks. SSRs have started to make an impact on speech tasks as well, with the first improvements seen in large-scale ASR with wav2vec~\citep{schneider2019wav2vec}. More recently, improvements have been seen on more tasks with wav2vec 2.0~\citep{baevski2020wav2vec} and other models~\citep{yang2021superb}.
% ~\citep{conneau2020unsupervised,hsu2021robust,chung2021w2v,hsu2021hubert,wu2021performance,ling2020decoar,yang2021superb}.

However, it is not yet clear how universal these pre-trained representations are for speech tasks, particularly for semantic understanding tasks like NER. Some pre-trained models achieve impressive performance across a variety of tasks, including some understanding tasks~\citep{yang2021superb} and analyses suggest that they contain at least some word meaning information~\citep{pasad2021layer}. However, these pre-trained models have not yet been tested on a broad range of challenging understanding tasks. We believe our work is the first to quantify the improvements from SSR, specifically on spoken NER.

\subsection{Leveraging external data}
{\it Self-training}~\citep{scudder1965probability,yarowsky1995unsupervised,riloff1996automatically} is a popular approach to improve supervised models when some additional unannotated data is available. Self-training has been observed to improve ASR~\citep{parthasarathi2019lessons, xu2021self} and is also complementary to pre-training~\citep{xu2021self}. To the best of our knowledge, this is the first work to introduce it to spoken NER while also studying its effects on both E2E and pipeline approaches. 

{\it Knowledge distillation} is widely used in model compression research. In this approach, some intermediate output from a teacher model is used to train a smaller student model~\citep{hinton2015distilling}. In the context of our work, the teacher and student networks are two different approaches to solving NER tasks, and the latter is trained on the final output tags of the former.

{\it Transfer learning} has been widely employed for SLU tasks~\citep{lugosch2019speech, jia2020large}, including E2E spoken NER~\citep{ ghannay2018end, caubriere2020we}. 
Automatic speech recognition (ASR) is a typically chosen task for pre-training a model before fine-tuning it for SLU. This choice is facilitated by the wider availability of transcribed speech than SLU annotations. Specifically for NER, ASR pre-training is expected to help since the accuracy of decoded texts can directly affect the final NER predictions.
\section{Methods}
\label{sec:approach}
Spoken NER involves detecting the entity phrases in a spoken utterance along with their tags. The annotations include the text transcripts for the audio and the entity phrases with their corresponding tags. Spoken NER, like any other SLU task, is typically tackled using one of two types of approaches: (i) Pipeline and (ii) End-to-end (E2E). As shown in Fig.~\ref{fig:baseline-systems}, a pipeline approach decodes speech to text using ASR and then passes the decoded text through a text NER module, whereas an E2E system directly maps the input speech to the output task labels. Each approach has its own set of advantages and shortcomings. Pipeline systems can enjoy the individual advances from both the speech and the text research communities, whereas combining two modules increases inference time, and propagation of ASR errors can have unexpected detrimental effects on the text NER module performance. On the other hand, E2E models directly optimize a task-specific objective and tend to have faster inference. However, such models typically require a large amount of task-specific labeled data to perform well. This can be seen from previous papers on E2E NER~\citep{yadav2020end, ghannay2018end}, where using at least 100 hours of labeled data is typical.

\subsection{Baseline models}
The baselines we use for E2E and pipeline models are taken from \citet{slue}. Similarly to previous work~\citep{slue, ghannay2018end, yadav2020end}, we formulate E2E NER as character-level prediction with tag-specific special characters delimiting entity phrases. For example, the phrases ``irish'' and ``eu'' are tagged as NORP\footnote{NORP: Nationalities or religious or political groups} (\textcolor{blue}{\$}) and GPE\footnote{GPE: Countries, cities, states} (\textcolor{blue}{\%}) respectively in ``\textit{the \textcolor{blue}{\$ irish ]} system works within a legal and regulatory policy directive framework dictated by the \textcolor{blue}{\% eu ]}}''.

The E2E NER and ASR modules are initialized with the wav2vec2.0 base~\citep{baevski2020wav2vec} pre-trained speech representation, while the text NER module is pre-trained with DeBERTa base~\citep{he2020deberta}. These pre-trained models are then fine-tuned for ASR/NER after adding a linear layer on top of the final hidden-state output. 

Since text transcripts are typically a part of the NER annotations, we can also train an NER model that uses the ground-truth text as input. This text NER model serves roughly as a topline and is further used in experiments with external data. The E2E NER and ASR models are trained with a character-level CTC objective. The text NER model is trained for token-level classification with cross-entropy loss. 

It is expected that using self-supervised representations gives a significant boost in limited labeled data settings. In order to quantify the benefits of the pre-trained representations in our setting, we also report the performance of E2E and pipeline baselines that are trained from scratch, not utilizing any pre-trained models.

\subsection{Evaluation metrics}
Similarly to previous work~\citep{ghannay2018end, yadav2020end}, we evaluate performance using {\it micro-averaged F1} scores on an unordered list of tuples of named entity phrase and tag pairs predicted for each sentence.
% F1 score is the harmonic mean of precision and recall.
An entity prediction is considered correct if both the entity text and the entity tag are correct.

Spoken NER introduces an added variability to the possible model errors due to speech-to-text conversion. We report word error rate (WER) to evaluate this aspect. {\it WER} is the word-level Levenshtein distance between the ground-truth text and the decoded text generated by the model. Additionally, to get an idea of the errors made by the model specifically on named entities, we also evaluate {\it NE ACC}, the proportion of entity phrases correctly decoded in the speech-to-text conversion. An entity phrase is considered accurate only if all the words in the phrase are correctly decoded in the right order. 

\subsection{Utilizing external data}
\label{sec:ext-data}
Next, we describe our approaches that use data external to the task-specific labeled data to improve both the pipeline and the E2E models for spoken NER. We consider four types of external data: (i) unlabeled speech ({\it Un-Sp}), (ii) unlabeled text ({\it Un-Txt}), (iii) transcribed speech ({\it Sp-Txt}), and (iv) text-based NER data.

% \begin{table*}[]
% \centering
% \begin{tabular}{ccccc}
% % \hlineB{2}      
% \toprule
% {\bf External data type} & {\bf Method}           & {\bf Labeling model} & {\bf Target model} & {\bf LM for decoding}            \\
% \midrule
% Unlabeled speech              & SelfTrain-ASR    & ASR            & ASR          & \multirow{3}{*}{T3 3-gram} \\ 
% Unlabeled text             & SelfTrain-txtNER & text NER       & text NER     &                            \\ 
% Transcribed speech             & Pre-ASR          & n/a            & n/a          &                            \\ 
% \bottomrule
% % \hlineB{2}
% \end{tabular}
% \caption{Methods for using external data for pipeline models. Details in Sec.~\ref{sec:ext-data}. \kl{I'm not quite sure that the "LM for decoding" column fits here, as it's not about how we use the external data.  I bold-faced the column headings in this table and the next.  I think this table (and the next) would look better with everything left-justified instead of centered.}}
% % \kl{say where all of the data/method/model/LM names are defined.}}
% \label{tab:approach-ppl}
% \end{table*}
\begin{table}[ht]
\centering
% \begin{tabular}{ccccc}
\begin{tabular}{llll}
% \hlineB{2}      
\toprule
\begin{tabular}[l]{@{}l@{}}{\bf External}\\ {\bf data type}\end{tabular} & {\bf Method}           & \begin{tabular}[l]{@{}l@{}}{\bf Target}\\ {\bf model}\end{tabular}             \\
\midrule
Un-Sp              & SelfTrain-ASR    &  ASR           \\ 
Un-Txt             & SelfTrain-txtNER &  text NER                                \\ 
Sp-Txt             & Pre-ASR          & ASR                                    \\ 
\bottomrule
% \hlineB{2}
\end{tabular}
\caption{Methods for using external data for pipeline models. For ``SelfTrain" approaches, the {\it labeling model} is the same as the {\it target model}. The method for external transcribed data ({\it Sp-Txt}) is based on transfer learning and thus there is no {\it labeling model}. More details are provided in Sec.~\ref{sec:ext-data}.}
% \kl{say where all of the data/method/model/LM names are defined.}}
\label{tab:approach-ppl}
\end{table}
% & \multirow{3}{*}{T3 3-gram}

\begin{table*}[ht]
\centering
% \begin{tabular}{ccccc}
\begin{tabular}{lllll}
% \hlineB{2}
\toprule
{\bf External data type}      & {\bf Method}           & {\bf Labeling model}                                                               & {\bf Target model} & {\bf LM for decoding} \\
% \hlineB{2}
\midrule
\multirow{2}{*}{Un-Sp}  & SelfTrain-E2E    & E2E-NER                                                                      & E2E-NER      & pLabel 4-gram   \\
                        & Distill-Pipeline & \begin{tabular}[l]{@{}l@{}}Pipeline-NER\\ (after SelfTrain-ASR)\end{tabular} & E2E-NER      & pLabel 4-gram   \\
\midrule
Un-Txt                  & Distill-txtNER-lm   & text NER                                                                     & n/a          & pLabel 4-gram   \\
\midrule
\multirow{2}{*}{Sp-Txt} & Distill-txtNER      & text NER                                                                     & E2E-NER      & pLabel 4-gram   \\
                        & Pre-ASR          & n/a                                                                          & n/a          & ftune 4-gram    \\ 
\bottomrule
% \hlineB{2}
\end{tabular}
\caption{Methods for using external data for E2E models. Details are provided in Sec.~\ref{sec:ext-data}.}
% \textbf{\kl{say where all of the data/method/model/LM names are defined.}}}
\label{tab:approach-e2e}
\end{table*}

The majority of techniques we consider involve labeling the external data with a {\it labeling model} (typically one of the baseline models) to produce {\it pseudo-labels}. The {\it target model} is then trained on these generated pseudo-labels along with the original labeled NER data. Tables~\ref{tab:approach-ppl} and \ref{tab:approach-e2e} present a detailed list of all methods we consider for improving pipeline and E2E models respectively. The methods we include use the first three kinds of external data listed above.  The fourth kind, external text-based NER data, is used in experiments attempting to improve the text NER model; since it does not succeed (Sec.~\ref{sec:ext-text-ner}), this data source is not explored further for the pipeline and E2E models.

When the labeling model is the same as the target model, this is a well-established process called self-training~\citep{scudder1965probability,yarowsky1995unsupervised,riloff1996automatically,xu2020iterative,xu2021self}. In our setting, a word-level language model (LM) is used for decoding both the ASR and E2E NER models. \citet{slue} observed that a LM consistently improves performance of all of the baseline models. So we may expect self-training from pseudo-labels to improve the target models by distilling the LM information into all model layers.

When the two models are different, we refer to it as knowledge distillation~\citep{hinton2015distilling}, where the information is being distilled from the labeling model to the target model. This approach enables the target model to learn from the better-performing labeling model via pseudo-labels. Among the baseline models, the pipeline performs better than E2E approaches, presumably since the former uses strong pre-trained text representations. So, for instance, distilling from the pipeline (labeling model) into the E2E model (target model) is expected to boost the performance of the E2E model.

The LMs used for decoding in different approaches are mentioned in Tab.~\ref{tab:approach-e2e}. All the ASR experiments use language models trained on the TED-LIUM 3 LM corpus~\citep{hernandez2018ted} as in \citet{slue}. The language model used in baseline E2E NER experiments is trained on the 15hr fine-tune set ({\it ftune 4-gram}). The generated pseudo-labels also provide additional annotated data for LM training, which can be used in E2E models. These are referred to as {\it plabel 4-gram}) (for "pseudo-label 4-gram"). 

\textbf{Unlabeled speech: } The unlabeled speech is used to improve the ASR module of the pipeline approach via self-training ({\it SelfTrain-ASR}).

For improving the E2E model, the improved pipeline can be used as the labeling model, followed by training the E2E model on the generated pseudo-labels ({\it Distill-Pipeline}). Alternatively, the unlabeled audio can be directly used to improve the E2E model via self-training ({\it SelfTrain-E2E}).

\textbf{Unlabeled text: } The text NER module in the pipeline approach is improved by self-training using the unlabeled text data ({\it SelfTrain-txtNER}). The E2E model uses the pseudo labels generated from the text NER baseline module on the unlabeled text to update the LM used for decoding ({\it Distill-txtNER-lm}).

\textbf{Transcribed speech: }The pipeline approach is improved by using the additional transcribed speech data to improve the ASR module ({\it Pre-ASR}). The E2E model uses this updated ASR as an initialization in a typical transfer learning setup.  Alternatively, for paired speech text data, the pseudo-labels generated from the text NER model can be paired with audio and used for training the E2E model, thus distilling information from a stronger text NER model into it ({\it Distill-txtNER}).

\textbf{Text NER data: } In addition to improving the pipeline and E2E models using the approaches mentioned above, we also look for any possible improvements in the text NER model by leveraging a larger external annotated text NER corpus. The DeBERTa base model is first fine-tuned on the larger external corpus, and then further fine-tuned on the in-domain labeled data. The first fine-tuning step is expected to help avoid shortcomings in performance due to the limited size of the in-domain labeled data. 

This approach is limited by the availability of external datasets with the same annotation scheme as the in-domain corpus. We use the OntoNotes5.0~\citep{pradhan2013towards} corpus, whose labeling scheme inspired that of VoxPopuli~\citep{slue}. See Tab.~\ref{tab:datastats} for more information on OntoNotes5.0.
\section{Experimental setup}
\label{sec:exp}
\subsection{Dataset}
\begin{table}[ht]
\centering
%\begin{tabular}{c|ccc}
% \small
\begin{tabular}{l|rrr}
%  \hlineB{2}
\toprule
%  \begin{tabular}[c]{@{}c@{}} \textbf{\# utter-}\\\textbf{ances} \end{tabular}
% \multirow{2}{*}{external}
% \
\textbf{Data split}   & \textbf{\# utt} & \begin{tabular}[c]{@{}c@{}} \textbf{Duration}\\\textbf{(hours)} \end{tabular} & \begin{tabular}[c]{@{}c@{}} \textbf{\# entity}\\\textbf{phrases} \end{tabular} \\
\midrule
fine-tune            & 5k              & 15    & 5820                  \\
dev                   & 1.7k            & 5    & 1862                 \\
test                  & 1.8k            & 5    & 2006      \\
\midrule
ext-100h &  350k            & 101         & \multirow{2}{*}{N/A}            \\
ext-500h & 177k            & 508                     \\
\begin{tabular}[l]{@{}l@{}}{ext-NER}\\{(ontonotes-}\\{train)}\end{tabular} & 66.6k & N/A & 81.8k\\

%  \hlineB{2}
\bottomrule
\end{tabular}
\caption{Data statistics. The ``ext-" prefix denotes external datasets. The external data doesn't have named entity annotations, except for OntoNotes 5.0.}
\label{tab:datastats}
\end{table}
VoxPopuli~\citep{wang2021voxpopuli} is a large multilingual speech corpus consisting of European Parliament event recordings with audio, transcripts, and timestamps from the official Parliament website. The English subset of the corpus has 540 hours of spoken data with text transcripts. \citet{slue} recently published NE annotations for a 15-hour subset of the train set and the complete standard dev set. Test set annotations are not public, but we obtain test set results by submitting model outputs following the SLUE site instructions.\footnote{\href{https://asappresearch.github.io/slue-toolkit/how-to-submit.html}{https://asappresearch.github.io/slue-toolkit/how-to-submit.html}} For our experiments with external in-domain data, we use uniformly sampled 100-hour and 500-hour subsets of the remainder of the VoxPopuli train set. The statistics for these splits are reported in Tab.~\ref{tab:datastats}. For more information on NE tags and label distribution, we direct the reader to the dataset and annotation papers~\citep{pradhan2013towards, slue}.

\subsection{Baseline models}
We closely follow the setup for E2E and pipeline baselines in \citet{slue}.\footnote{\href{https://github.com/asappresearch/slue-toolkit/blob/main/README.md}{https://github.com/asappresearch/slue-toolkit/blob/main/README.md}}
We use wav2vec 2.0 base~\citep{baevski2020wav2vec} and DeBERTa-base~\citep{he2020deberta} as the unsupervised pre-trained models, which have 95M and 139M parameters respectively. The publicly available wav2vec2.0 base model is pre-trained on 960 hours of the LibriSpeech audiobooks corpus~\citep{panayotov2015librispeech}.

For baselines that do not use pre-trained representations, we utilize the DeepSpeech2 (DS2) toolkit\footnote{\href{https://github.com/SeanNaren/deepspeech.pytorch}{https://github.com/SeanNaren/deepspeech.pytorch}}~\citep{amodei2016deep}. DS2 first converts audio files into spectrogram features. The model processes the spectrogram features through two 2-D convolutional layers followed by five bidirectional 2048-dim LSTM layers and a softmax layer. The softmax layer outputs the probabilities for a sequence of characters. The model has 26M parameters and is trained with SpecAugment data augmentation~\citep{park2019specaugment} and a character-level CTC objective. Following~\citet{slue}, we train on the finer label set (18 entity tags) and evaluate on the combined version (7 entity tags).

\subsection{Utilizing external data}
We use fairseq library~\citep{ott2019fairseq} to fine-tune wav2vec 2.0 models for the E2E NER and ASR tasks. We fine-tune the model for 80k (160k) updates on 100 (500) hours of pseudo-labeled data. It takes 20 (40) hours (wall clock time) to fine-tune on 100 (500) hours of data using 8 TITAN RTX GPUs. We use HuggingFace's transformers toolkit~\citep{wolf2019huggingface} for training the text NER model on pseudo-labels. Detailed config files are provided in our codebase.\footnote{\href{https://github.com/asappresearch/spoken-ner}{https://github.com/asappresearch/spoken-ner}}

\section{Results}
\label{sec:results}
% \subsection{Effect of self-supervised pre-training}
% \input{tables/ssl}
% Results in Table \ref{tab:ssl}. We notice that w2v2-ls model outperforms w2v2-multi-100k model suggesting that language mismatch hurts the performance more than domain mismatch does. Seeing that w2v2-ls performs the best, we perform remainder of the experiments using w2v2-ls as the pre-trained model.
\subsection{Baseline models}
% \begin{table}[]
% \centering
% \begin{tabular}{cc|cc}
% \hlineB{2}
% Approach & \begin{tabular}[c]{@{}c@{}}Pretrained\\ model\end{tabular} & F1  & Label-F1 \\
% \hlineB{2}
% Pipeline & N/A & \tba & \tba \\
% E2E & N/A & 43.9  & 51.8      \\ \hline
% Pipeline & W2V2-B & 64.7 & 77.1     \\
% NLP Topline  & DeBERTa-B & 81.4 & 86.3      \\
% E2E & W2V2-B & 63.4 & 71.7  \\
% \hlineB{2}
% \end{tabular}

\begin{table}[h]
\centering
\begin{tabular}{l|cc|c}
% \hlineB{2}
\toprule
\multirow{2}{*}{\begin{tabular}[l]{@{}l@{}}{\bf NER}\\{\bf system}\end{tabular}} & \multicolumn{2}{c|}{\bf Pretrained model} & \multirow{2}{*}{\bf F1} \\
 & {\bf Speech} & {\bf Text} & \\
\midrule
% \hlineB{2}
Pipeline & \xmark & DeBERTa-B & 52.4 \\
E2E & \xmark & \xmark & 51.8       \\
\midrule
Pipeline & W2V2-B & DeBERTa-B & 72.0      \\
E2E & W2V2-B & \xmark & 68.1  \\
\midrule
Text NER & \xmark & DeBERTa-B & 86.0      \\
\bottomrule
% \hlineB{2}
\end{tabular}
\caption{Dev set \% f-score performance of baseline models. All models here are trained on the 15hr fine-tune set. The pre-trained speech and text models are mentioned wherever used or applicable. The last three rows are from previously established baselines~\citep{slue}.}
\label{tab:baseline}
\end{table}
Results from all the baseline models are reported in Tab.~\ref{tab:baseline}. The models here are trained on the 15hr fine-tune set. We see that self-supervised pre-training gives a significant performance boost over no pre-training. The text NER model (which uses ground-truth transcripts) is far better than the pipeline method, which is better than the E2E model.

\subsection{Leveraging external data}
\begin{table}[h]
\centering
\begin{tabular}{l|l|cc}
% \hlineB{2}
\toprule
{\bf Ext. data} & {\bf Method}           & \multicolumn{1}{c}{{\bf 100h}} & \multicolumn{1}{c}{{\bf 500h}} \\
\midrule
Un-Sp              & SelfTrain-ASR    &       73.8                   &       74.4                   \\
Un-Txt             & SelfTrain-txtNER &         72.3                 &         70.8                 \\
Sp-Txt             & Pre-ASR          &       75.6                   &       77.7                  \\
\bottomrule
\end{tabular}
\caption{Dev set \% f-score performance of the pipeline models. Note the baseline Pipeline (72) and text NER (86.0) performances without using any additional data from Tab.~\ref{tab:baseline}.}
\label{tab:results-ppl}
% \vspace{-20pt}
\end{table}
% \usepackage{multirow}
\begin{table}[h]
\centering
\begin{tabular}{l|l|cc}
% \hlineB{2}
\toprule
{\bf Ext. data}      & {\bf Method}           & \multicolumn{1}{c}{{\bf 100h}} & \multicolumn{1}{c}{{\bf 500h}} \\
% \midrule
% None & Baseline E2E & \multicolumn{2}{c}{68.1} \\
\midrule
\multirow{2}{*}{Un-Sp}  & SelfTrain-E2E    &          70.6                &       72.1                   \\
                        & Distill-Pipeline &          76.5                &    77.5                      \\
\midrule
Un-Txt                  & Distill-txtNER-lm      &         71.0                 &         71.7                 \\
\midrule
\multirow{2}{*}{Sp-Txt} & Distill-txtNER      &       79.2                   &          82.2               \\
                        & Pre-ASR          &         70.7                 &       73.2                  \\
% \midrule
% None & Text NER & \multicolumn{2}{c}{86.0} \\
\bottomrule
\end{tabular}
\caption{Dev set \% f-score performance of the E2E models. Note the baseline E2E (68.1) and text NER (86.0) performances without using any additional data from Tab.~\ref{tab:baseline}.}
\label{tab:results-e2e}
% \vspace{-20pt}
\end{table}
We report F1 scores on the dev set using different pipeline and E2E approaches in Tables~\ref{tab:results-ppl} and \ref{tab:results-e2e} respectively. Fig.~\ref{fig:summary} presents key results when using each external data type for both E2E and pipeline models. The key findings are: \\
(i) Using external data reduces the gap between spoken NER baselines and text NER. \\
(ii) With access to either unlabeled speech or transcribed speech, E2E models outperform pipeline models, whereas, for the baselines, the opposite holds. \\
(iii) Using unlabeled text gives the smallest boost among the three types of external data, and the pipeline approach performs better in that setting.

A summary of test set results is presented in Appendix~\ref{app:test-results}. The results follow the same trend as on the dev set.

\subsubsection{External text NER data}
\label{sec:ext-text-ner}
We try to improve the text NER model by using the OntoNotes5.0 NER corpus~\citep{pradhan2013towards}. Fine-tuning DeBERTa-base on OntoNotes5.0 produces an F1 of 60\% on the VoxPopuli dev set. Fine-tuning it further on VoxPopuli gives F1 86\% on the dev set. Since we do not see any boost over the existing vanilla approach (86\%, see Tab.~\ref{tab:baseline}), we retain the original text NER model using only in-domain data and do not perform further experiments using the OntoNotes-finetuned model. 

\section{Discussion and analysis}
The baseline results are not surprising:  The limited labeled data is not enough for the baseline E2E approach, but the pipeline model can leverage a strong text representation model, which gives it an edge. Improvements to these models can be attributed to either (i) a better speech-to-text conversion or (ii) a better semantic understanding of the input content. Next, we use this distinction to understand the observed improvements from using external data.

\subsection{Improved E2E results}
When using external data with the E2E model, the best performing methods use either (a) external unlabeled speech ({\it Distill-Pipeline}) or (b) transcribed speech ({\it Distill-txtNER}). The labeling models have a stronger semantic component than the E2E baseline in both of these scenarios because of their strong text NER module. The same cannot be said for the other competing approaches for these external data categories, {\it SelfTrain-NER} and {\it Pre-ASR}, which provide much lower improvements. {\it SelfTrain-NER} distills information from the LM into the model layers, but the n-gram LM is much less powerful than the transformer-based text NER module used in {\it Distill-Pipeline}. The {\it Pre-ASR} approach has no means to improve the semantic component in the updated model.

In the presence of unlabeled text data, the modification comes from a better 4-gram LM trained on pseudo-labels. Note that the baseline E2E model parameters do not change, unlike when using the other two types of external data. This can explain why this approach only has a small improvement over the baseline.

\subsection{Improved pipeline results}
The baseline pipeline model already takes advantage of the text NER module, which leaves little room for improvement in the semantic understanding component. Specifically, using unlabeled text data to improve the text NER module ({\it SelfTrain-txtNER}) gives a small boost of 0.4\%. For comparison, note that the improvement from using unlabeled speech is 2.5\% over baseline. So, the hope with pipeline models is for the external data to improve the speech-to-text conversion, which can then help reduce error propagation between the independent pipeline modules.

\subsection{Amount of external data}
Almost all experiments produce a larger improvement when using 500 hours of external data than 100 hours. Only {\it SelfTrain-txtNER} has a reverse trend (see Tab.~\ref{tab:results-ppl}). The higher amount of external data naturally increases the fraction of noisy data that the target model is trained on, and that may lead to a poorer model. We hypothesize that methods for balancing between the effects of manually annotated and pseudo-labeled examples could help tackle this issue~\citep{park2020improved}. However, we leave an in-depth investigation of this phenomenon to future work.

\subsection{Error analysis}
\label{sec:analysis}
For analysis, we choose the best-performing models within each category.
\begin{figure}[h]
% {<left> <lower> <right> <upper>}
\begin{minipage}[b]{1.0\linewidth}
\small
% \vspace{-0.5cm}
 \centering
 \centerline{\includegraphics[width=8cm, trim=0 105 0 125, clip]{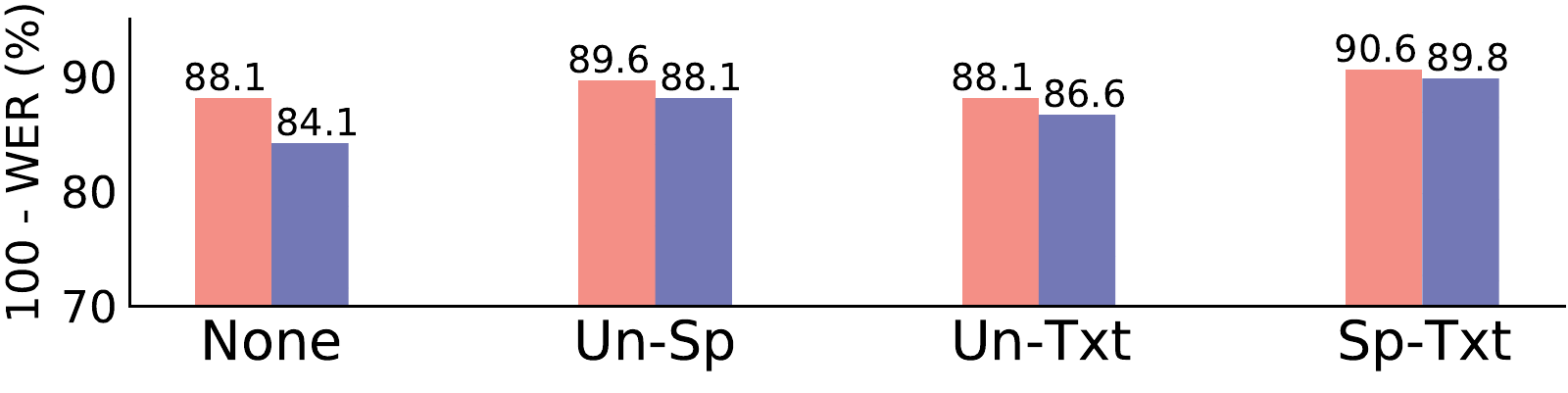}}
\end{minipage}
\begin{minipage}[b]{1.0\linewidth}

\vspace{-0.05cm}
\footnotesize
 \centering
  \centerline{\includegraphics[width=8cm, trim=0 85 0 95, clip]{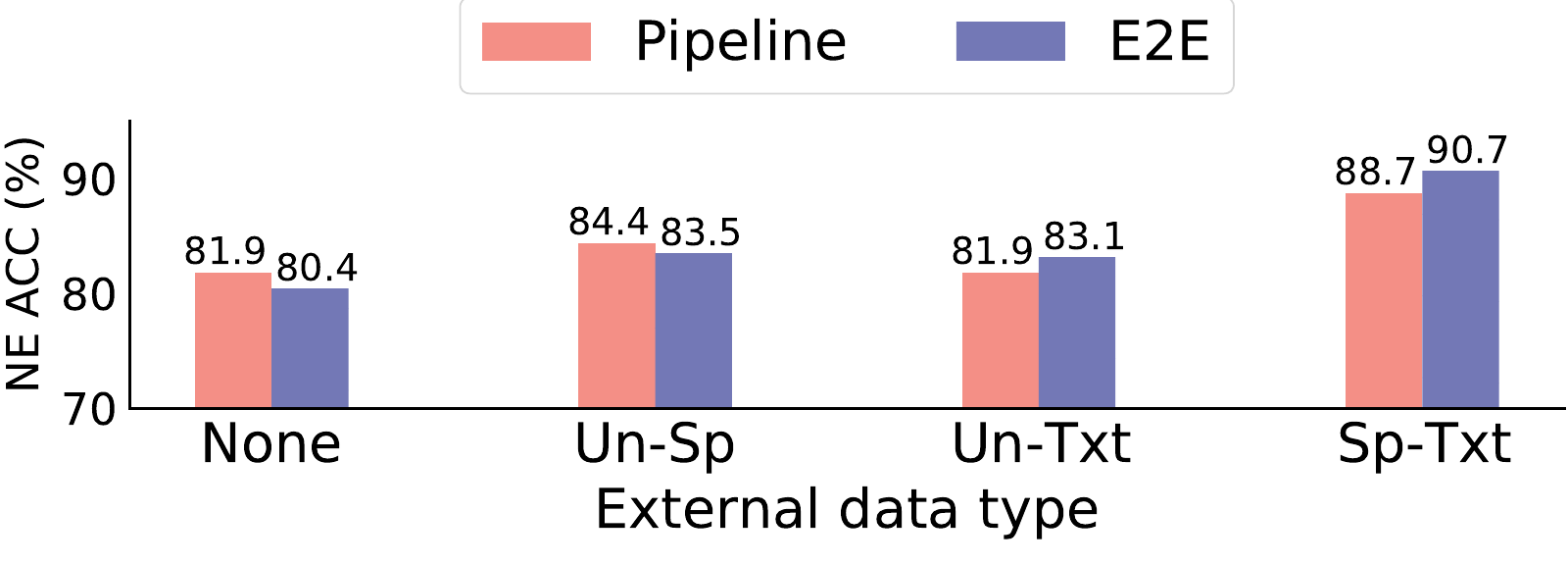}}
\end{minipage}

\vspace{-0.3cm}
\caption{{\it $100-$WER (\%) and NE ACC (\%) values on the dev set for the best-performing models in each category with access to 100 hours of external data.}}
  \label{fig:wer-ne-acc}
\vspace{-0.3cm}
\end{figure}
\begin{figure}[h]
% {<left> <lower> <right> <upper>}
\begin{minipage}[b]{1.0\linewidth}
\small
% \vspace{-0.5cm}
 \centering
 \centerline{\includegraphics[width=8cm, trim=0 75 0 100, clip]{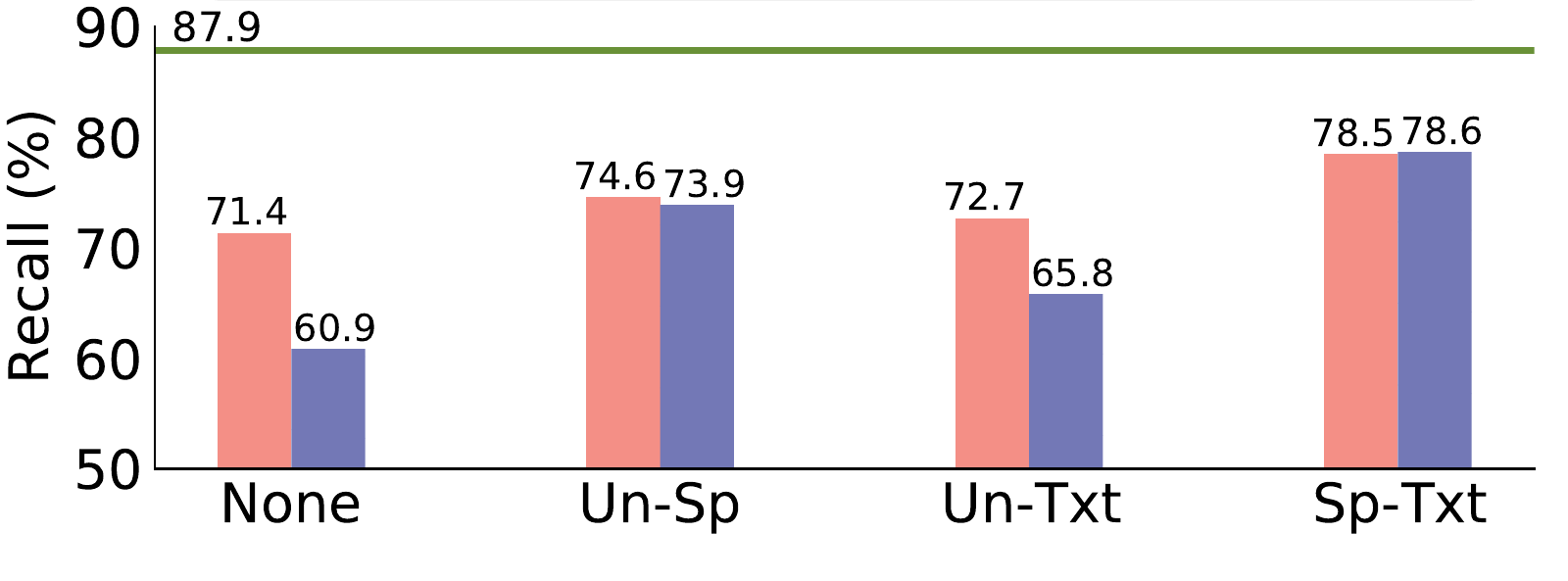}}
\end{minipage}
\begin{minipage}[b]{1.0\linewidth}

\vspace{-0.05cm}
\footnotesize
 \centering
  \centerline{\includegraphics[width=8cm, trim=0 65 0 75, clip]{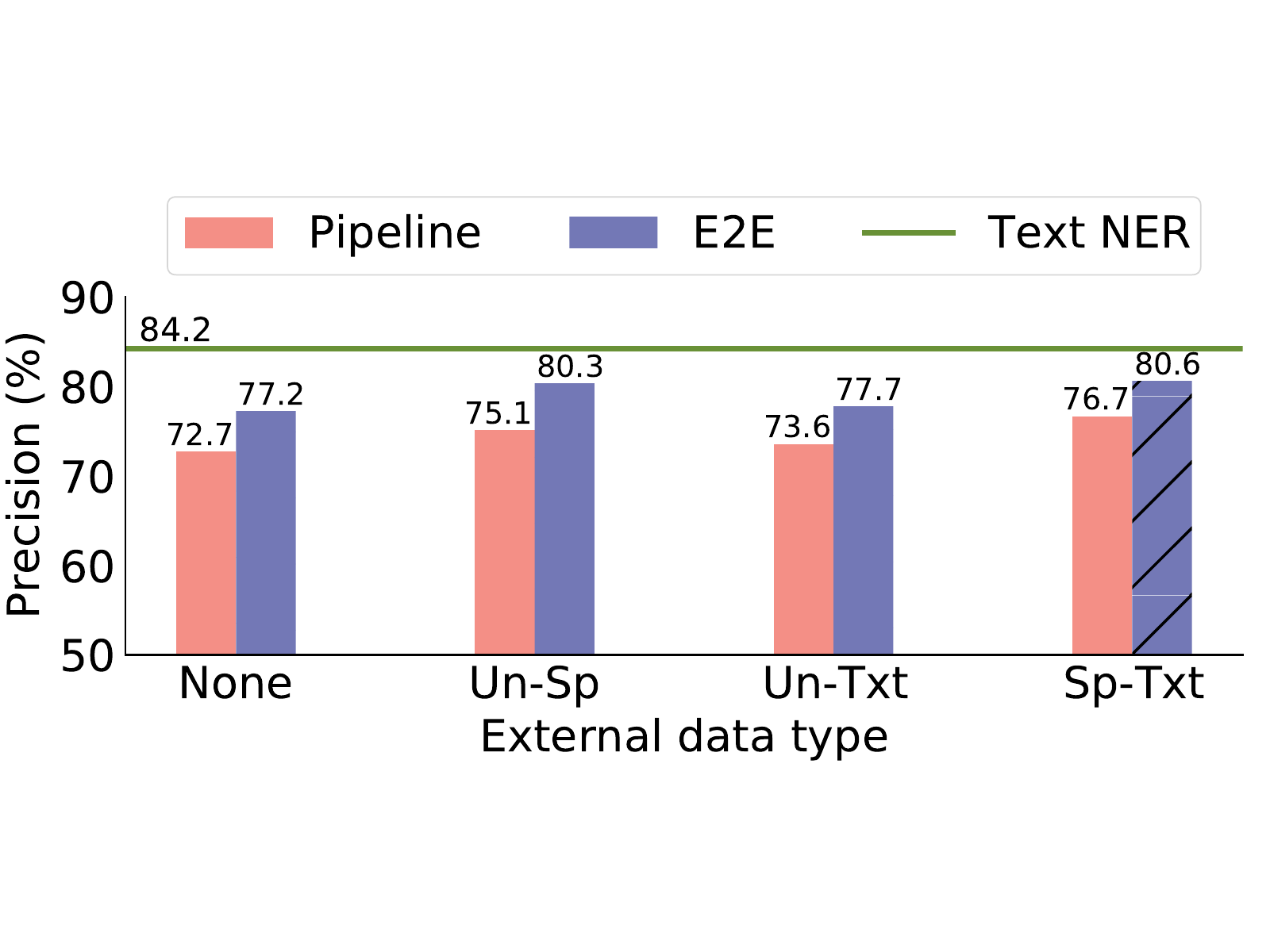}}
\end{minipage}

\vspace{-0.3cm}
\caption{{\it Recall and precision on the dev set for the best-performing models in each category with access to 100 hours of external data.}}
  \label{fig:prec-recall}
\vspace{-0.3cm}
\end{figure}

Fig.~\ref{fig:wer-ne-acc} presents the NE accuracy and word error rates (WER). We strip off the tag-specific special character tokens when evaluating WER for the E2E NER models. Note that we report $100 - $ WER so that higher is better in both plots. We observe that the ASR used in pipeline models typically performs better than the speech-to-text conversion of E2E models, even when the former has a poorer F1 (Fig.~\ref{fig:summary}). This may lead us to hypothesize that the E2E model recognizes NE words better while doing worse for other words. However, this hypothesis is not supported by the {\it NE-ACC} results (Fig.~\ref{fig:wer-ne-acc}).

Next, we look at the breakdown of F1 into precision and recall (Fig.~\ref{fig:prec-recall}). We see that pipeline models have worse precision, thus suggesting that these suffer from a higher false-positive rate than the E2E models. This explains why {\it NE-ACC} is not predictive of F1; the former can inform us about errors due to false negatives, but not false positives.

\subsubsection{Error categories}
\begin{figure*}[h]
    \centering
    \hspace{-.15in}\includegraphics[width=\linewidth]{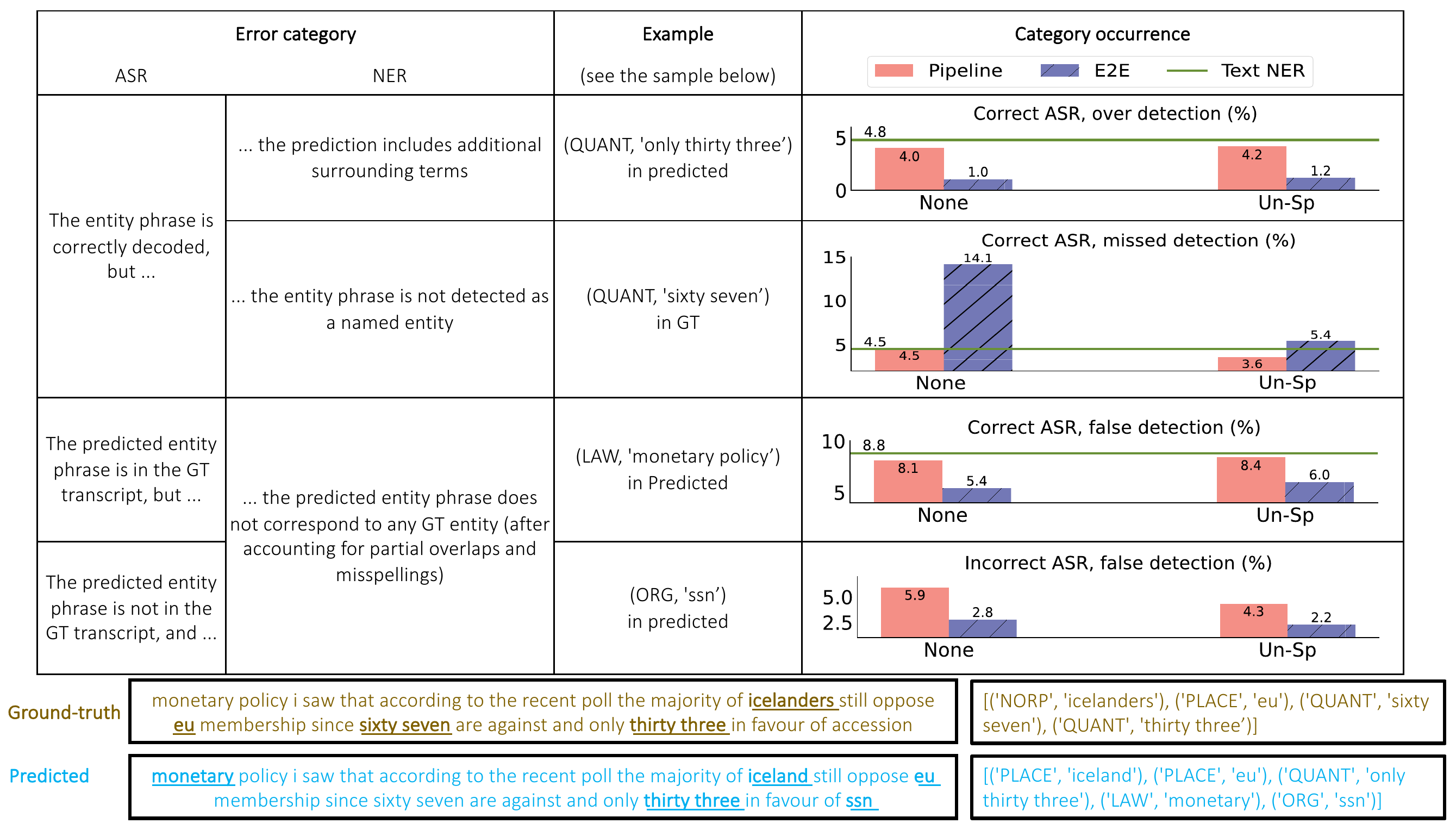}
    \caption{\it NER error category distribution on the dev set. The category-specific error rates in the plots are normalized by the total number of ground-truth (GT) entities.
    % \klremove{ when the models are evaluated on the dev set}.
    The examples here are artificially created from the same ground-truth example for ease of presentation. Actual examples of these categories are presented in Appendix~\ref{app:error-cat}.
    % \kl{time permitting it would be great to have the bar plots line up (i.e. so that the "None" and "Un-Sp" bars are all on top of each other.}
    }
    \label{fig:error-cat}
\end{figure*}
For a more detailed understanding of model behavior, we categorize the NER errors into an exhaustive list of types (details in Appendix~\ref{app:error-cat}). We focus on four major categories showing noteworthy differences between pipeline and E2E approaches. We provide this analysis for the baselines, {\it Distill-Pipeline}, and {\it SelfTrain-ASR} models using external unlabeled speech data. The trends and observations presented here are consistent with the other two external data types. 

The major error categories, along with examples, are presented in Fig.~\ref{fig:error-cat}. We observe that:\\
(i) False detections are 1.5 times more common in pipeline models than in E2E models, as expected based on the lower precision for the former. This happens even when the falsely detected text is not a speech-to-text conversion error.\\
(ii) Over-detections are 3.5 to 4 times more common in the pipeline models even when the entity phrase is decoded correctly.\\
(iii) Missed detections for the E2E {\it Distill-Pipeline} model are drastically reduced compared to the E2E baseline. Missed detections refer to cases where the entity phrases are correctly transcribed but are not labeled as named entities.  The improvement here therefore suggests that {\it Distill-Pipeline} improves the understanding capability of the E2E model, in addition to its speech-to-text capability. Also, note that the pipeline model does not enjoy the same benefit from unlabeled speech since this only involves self-training (instead of knowledge distillation from a much richer model for E2E).

Overall, the pipeline models suffer disproportionately from false positives. This seems to stem from the text NER model, which has even higher over-detection and false detection rates than the pipeline baseline models (Fig.~\ref{fig:error-cat}). The reasons behind this difference between E2E and pipeline models need further investigation.

\section{Conclusion}
We have explored various ways to use different external data types that improve both pipeline and E2E methods for spoken NER. The best-performing model when using external data is an E2E approach. This is one of the few results in the literature thus far showing better performance for E2E over pipeline methods that use state-of-the-art modules for spoken language understanding. We develop some insights into this difference; we notice that pipeline models are adversely affected by false positives and that leveraging external data improves the semantic understanding capability of the E2E models.

We hope that our work provides guiding principles for researchers working on SLU tasks in similar low-resource domains when some form of external data is abundant. This work also leaves some interesting research questions for future work. For example, we see minor improvements between 100h and 500h of external data (see Tab.~\ref{tab:results-ppl} and \ref{tab:results-e2e}), which suggests the question: What is the smallest amount of external data needed to obtain significant improvements in NER performance? Additionally, one preliminary experiment with external, out-of-domain text NER data (OntoNotes 5.0) fails to improve the text NER performance, suggesting the challenges of dealing with out-of-domain datasets. The scenario where we have access to out-of-domain external data is common but challenging, and warrants further study. From the modeling perspective, better fine-tuning strategies for wav2vec2.0 in low supervision settings have been proposed for ASR~\citep{pasad2021layer}; it would be interesting to explore how these findings may transfer to an SLU task.

\section*{Acknowledgements}
We thank Yoav Artzi for his feedback and inputs to strengthen our experimental setup and analysis.  We also thank the anonymous reviewers for the detailed comments and suggestions.

\bibliography{refs}

\begin{thebibliography}{44}
\expandafter\ifx\csname natexlab\endcsname\relax\def\natexlab#1{#1}\fi

\bibitem[{Amodei et~al.(2016)Amodei, Ananthanarayanan, Anubhai, Bai,
  Battenberg, Case, Casper, Catanzaro, Chen, Chrzanowski, Coates, Diamos,
  Elsen, Engel, Fan, Fougner, Hannun, Jun, Han, LeGresley, Li, Lin, Narang, Ng,
  Ozair, Prenger, Qian, Raiman, Satheesh, Seetapun, Sengupta, Wang, Wang, Wang,
  Xiao, Xie, Yogatama, Zhan, and Zhu}]{amodei2016deep}
Dario Amodei, Sundaram Ananthanarayanan, Rishita Anubhai, Jingliang Bai, Eric
  Battenberg, Carl Case, Jared Casper, Bryan Catanzaro, Jingdong Chen, Mike
  Chrzanowski, Adam Coates, Greg Diamos, Erich Elsen, Jesse~H. Engel, Linxi
  Fan, Christopher Fougner, Awni~Y. Hannun, Billy Jun, Tony Han, Patrick
  LeGresley, Xiangang Li, Libby Lin, Sharan Narang, Andrew~Y. Ng, Sherjil
  Ozair, Ryan Prenger, Sheng Qian, Jonathan Raiman, Sanjeev Satheesh, David
  Seetapun, Shubho Sengupta, Chong Wang, Yi~Wang, Zhiqian Wang, Bo~Xiao, Yan
  Xie, Dani Yogatama, Jun Zhan, and Zhenyao Zhu. 2016.
\newblock \href {http://proceedings.mlr.press/v48/amodei16.html} {Deep speech 2
  : End-to-end speech recognition in english and mandarin}.
\newblock In \emph{Proceedings of the 33nd International Conference on Machine
  Learning, {ICML} 2016, New York City, NY, USA, June 19-24, 2016}, volume~48
  of \emph{{JMLR} Workshop and Conference Proceedings}, pages 173--182.
  JMLR.org.

\bibitem[{Baevski et~al.(2020)Baevski, Zhou, Mohamed, and
  Auli}]{baevski2020wav2vec}
Alexei Baevski, Yuhao Zhou, Abdelrahman Mohamed, and Michael Auli. 2020.
\newblock \href
  {https://proceedings.neurips.cc/paper/2020/hash/92d1e1eb1cd6f9fba3227870bb6d7f07-Abstract.html}
  {wav2vec 2.0: {A} framework for self-supervised learning of speech
  representations}.
\newblock In \emph{Advances in Neural Information Processing Systems 33: Annual
  Conference on Neural Information Processing Systems 2020, NeurIPS 2020,
  December 6-12, 2020, virtual}.

\bibitem[{Bastianelli et~al.(2020)Bastianelli, Vanzo, Swietojanski, and
  Rieser}]{bastianelli2020slurp}
Emanuele Bastianelli, Andrea Vanzo, Pawel Swietojanski, and Verena Rieser.
  2020.
\newblock \href {https://doi.org/10.18653/v1/2020.emnlp-main.588} {{SLURP}: A
  spoken language understanding resource package}.
\newblock In \emph{Proceedings of the 2020 Conference on Empirical Methods in
  Natural Language Processing (EMNLP)}, pages 7252--7262, Online. Association
  for Computational Linguistics.

\bibitem[{Borgholt et~al.(2021)Borgholt, Havtorn, Abdou, Edin, Maal{\o}e,
  S{\o}gaard, and Igel}]{borgholt2021we}
Lasse Borgholt, Jakob~Drachmann Havtorn, Mostafa Abdou, Joakim Edin, Lars
  Maal{\o}e, Anders S{\o}gaard, and Christian Igel. 2021.
\newblock \href {https://arxiv.org/abs/2111.14842} {Do we still need automatic
  speech recognition for spoken language understanding?}
\newblock \emph{ArXiv preprint}, abs/2111.14842.

\bibitem[{Caubri{\`e}re et~al.(2020)Caubri{\`e}re, Rosset, Est{\`e}ve, Laurent,
  and Morin}]{caubriere2020we}
Antoine Caubri{\`e}re, Sophie Rosset, Yannick Est{\`e}ve, Antoine Laurent, and
  Emmanuel Morin. 2020.
\newblock \href {https://aclanthology.org/2020.lrec-1.556} {Where are we in
  named entity recognition from speech?}
\newblock In \emph{Proceedings of the 12th Language Resources and Evaluation
  Conference}, pages 4514--4520, Marseille, France. European Language Resources
  Association.

\bibitem[{Chen et~al.(2017)Chen, Fisch, Weston, and Bordes}]{chen2017reading}
Danqi Chen, Adam Fisch, Jason Weston, and Antoine Bordes. 2017.
\newblock \href {https://doi.org/10.18653/v1/P17-1171} {Reading {W}ikipedia to
  answer open-domain questions}.
\newblock In \emph{Proceedings of the 55th Annual Meeting of the Association
  for Computational Linguistics (Volume 1: Long Papers)}, pages 1870--1879,
  Vancouver, Canada. Association for Computational Linguistics.

\bibitem[{Devlin et~al.(2019)Devlin, Chang, Lee, and
  Toutanova}]{devlin2018bert}
Jacob Devlin, Ming-Wei Chang, Kenton Lee, and Kristina Toutanova. 2019.
\newblock \href {https://doi.org/10.18653/v1/N19-1423} {{BERT}: Pre-training of
  deep bidirectional transformers for language understanding}.
\newblock In \emph{Proceedings of the 2019 Conference of the North {A}merican
  Chapter of the Association for Computational Linguistics: Human Language
  Technologies, Volume 1 (Long and Short Papers)}, pages 4171--4186,
  Minneapolis, Minnesota. Association for Computational Linguistics.

\bibitem[{Ghannay et~al.(2018)Ghannay, Caubri{\`e}re, Est{\`e}ve, Camelin,
  Simonnet, Laurent, and Morin}]{ghannay2018end}
Sahar Ghannay, Antoine Caubri{\`e}re, Yannick Est{\`e}ve, Nathalie Camelin,
  Edwin Simonnet, Antoine Laurent, and Emmanuel Morin. 2018.
\newblock End-to-end named entity and semantic concept extraction from speech.
\newblock In \emph{SLT}.

\bibitem[{Graves et~al.(2006)Graves, Fern{\'{a}}ndez, Gomez, and
  Schmidhuber}]{graves2006connectionist}
Alex Graves, Santiago Fern{\'{a}}ndez, Faustino~J. Gomez, and J{\"{u}}rgen
  Schmidhuber. 2006.
\newblock \href {https://doi.org/10.1145/1143844.1143891} {Connectionist
  temporal classification: labelling unsegmented sequence data with recurrent
  neural networks}.
\newblock In \emph{Machine Learning, Proceedings of the Twenty-Third
  International Conference {(ICML} 2006), Pittsburgh, Pennsylvania, USA, June
  25-29, 2006}, volume 148 of \emph{{ACM} International Conference Proceeding
  Series}, pages 369--376. {ACM}.

\bibitem[{He et~al.(2021)He, Liu, Gao, and Chen}]{he2020deberta}
Pengcheng He, Xiaodong Liu, Jianfeng Gao, and Weizhu Chen. 2021.
\newblock \href {https://openreview.net/forum?id=XPZIaotutsD} {Deberta:
  decoding-enhanced bert with disentangled attention}.
\newblock In \emph{9th International Conference on Learning Representations,
  {ICLR} 2021, Virtual Event, Austria, May 3-7, 2021}. OpenReview.net.

\bibitem[{Hernandez et~al.(2018)Hernandez, Nguyen, Ghannay, Tomashenko, and
  Est{\`{e}}ve}]{hernandez2018ted}
Fran{\c{c}}ois Hernandez, Vincent Nguyen, Sahar Ghannay, Natalia Tomashenko,
  and Yannick Est{\`{e}}ve. 2018.
\newblock {TED-LIUM 3: Twice as Much Data and Corpus Repartition for
  Experiments on Speaker Adaptation}.
\newblock In \emph{SPECOM}.

\bibitem[{Hinton et~al.(2014)Hinton, Vinyals, and Dean}]{hinton2015distilling}
Geoffrey Hinton, Oriol Vinyals, and Jeff Dean. 2014.
\newblock Distilling the knowledge in a neural network.
\newblock In \emph{NIPS Deep Learning Workshop}.

\bibitem[{Jannet et~al.(2015)Jannet, Galibert, Adda-Decker, and
  Rosset}]{jannet2015evaluate}
Mohamed Ameur~Ben Jannet, Olivier Galibert, Martine Adda-Decker, and Sophie
  Rosset. 2015.
\newblock How to evaluate {ASR} output for named entity recognition?
\newblock In \emph{Interspeech}.

\bibitem[{Jia et~al.(2020)Jia, Wang, Zhang, Cheng, and Xiao}]{jia2020large}
Xueli Jia, Jianzong Wang, Zhiyong Zhang, Ning Cheng, and Jing Xiao. 2020.
\newblock \href {https://doi.org/10.21437/Interspeech.2020-0059} {Large-scale
  transfer learning for low-resource spoken language understanding}.
\newblock In \emph{Interspeech 2020, 21st Annual Conference of the
  International Speech Communication Association, Virtual Event, Shanghai,
  China, 25-29 October 2020}, pages 1555--1559. {ISCA}.

\bibitem[{Kim and Woodland(2000)}]{kim2000rule}
Ji-Hwan Kim and Philip~C Woodland. 2000.
\newblock A rule-based named entity recognition system for speech input.
\newblock In \emph{ICSLP}.

\bibitem[{Li et~al.(2020{\natexlab{a}})Li, Sun, Han, and Li}]{li2020survey}
Jing Li, Aixin Sun, Jianglei Han, and Chenliang Li. 2020{\natexlab{a}}.
\newblock A survey on deep learning for named entity recognition.
\newblock \emph{IEEE Transactions on Knowledge and Data Engineering}.

\bibitem[{Li et~al.(2020{\natexlab{b}})Li, Sun, Meng, Liang, Wu, and
  Li}]{li2019dice}
Xiaoya Li, Xiaofei Sun, Yuxian Meng, Junjun Liang, Fei Wu, and Jiwei Li.
  2020{\natexlab{b}}.
\newblock \href {https://doi.org/10.18653/v1/2020.acl-main.45} {Dice loss for
  data-imbalanced {NLP} tasks}.
\newblock In \emph{Proceedings of the 58th Annual Meeting of the Association
  for Computational Linguistics}, pages 465--476, Online. Association for
  Computational Linguistics.

\bibitem[{Louvan and Magnini(2018)}]{louvan-magnini-2018-exploring}
Samuel Louvan and Bernardo Magnini. 2018.
\newblock \href {https://doi.org/10.18653/v1/W18-5711} {Exploring named entity
  recognition as an auxiliary task for slot filling in conversational language
  understanding}.
\newblock In \emph{Proceedings of the 2018 {EMNLP} Workshop {SCAI}: The 2nd
  International Workshop on Search-Oriented Conversational {AI}}, pages 74--80,
  Brussels, Belgium. Association for Computational Linguistics.

\bibitem[{Lugosch et~al.(2019)Lugosch, Ravanelli, Ignoto, Tomar, and
  Bengio}]{lugosch2019speech}
Loren Lugosch, Mirco Ravanelli, Patrick Ignoto, Vikrant~Singh Tomar, and Yoshua
  Bengio. 2019.
\newblock \href {https://doi.org/10.21437/Interspeech.2019-2396} {Speech model
  pre-training for end-to-end spoken language understanding}.
\newblock In \emph{Interspeech 2019, 20th Annual Conference of the
  International Speech Communication Association, Graz, Austria, 15-19
  September 2019}, pages 814--818. {ISCA}.

\bibitem[{Nadeau and Sekine(2007)}]{nadeau2007survey}
David Nadeau and Satoshi Sekine. 2007.
\newblock A survey of named entity recognition and classification.
\newblock \emph{Lingvisticae Investigationes}.

\bibitem[{Ott et~al.(2019)Ott, Edunov, Baevski, Fan, Gross, Ng, Grangier, and
  Auli}]{ott2019fairseq}
Myle Ott, Sergey Edunov, Alexei Baevski, Angela Fan, Sam Gross, Nathan Ng,
  David Grangier, and Michael Auli. 2019.
\newblock \href {https://doi.org/10.18653/v1/N19-4009} {fairseq: A fast,
  extensible toolkit for sequence modeling}.
\newblock In \emph{Proceedings of the 2019 Conference of the North {A}merican
  Chapter of the Association for Computational Linguistics (Demonstrations)},
  pages 48--53, Minneapolis, Minnesota. Association for Computational
  Linguistics.

\bibitem[{Panayotov et~al.(2015)Panayotov, Chen, Povey, and
  Khudanpur}]{panayotov2015librispeech}
Vassil Panayotov, Guoguo Chen, Daniel Povey, and Sanjeev Khudanpur. 2015.
\newblock \href {https://doi.org/10.1109/ICASSP.2015.7178964} {Librispeech: An
  {ASR} corpus based on public domain audio books}.
\newblock In \emph{2015 {IEEE} International Conference on Acoustics, Speech
  and Signal Processing, {ICASSP} 2015, South Brisbane, Queensland, Australia,
  April 19-24, 2015}, pages 5206--5210. {IEEE}.

\bibitem[{Parada et~al.(2011)Parada, Dredze, and Jelinek}]{parada2011oov}
Carolina Parada, Mark Dredze, and Frederick Jelinek. 2011.
\newblock {OOV} sensitive named-entity recognition in speech.
\newblock In \emph{Interspeech}.

\bibitem[{Park et~al.(2019)Park, Chan, Zhang, Chiu, Zoph, Cubuk, and
  Le}]{park2019specaugment}
Daniel~S. Park, William Chan, Yu~Zhang, Chung{-}Cheng Chiu, Barret Zoph,
  Ekin~D. Cubuk, and Quoc~V. Le. 2019.
\newblock \href {https://doi.org/10.21437/Interspeech.2019-2680} {Specaugment:
  {A} simple data augmentation method for automatic speech recognition}.
\newblock In \emph{Interspeech 2019, 20th Annual Conference of the
  International Speech Communication Association, Graz, Austria, 15-19
  September 2019}, pages 2613--2617. {ISCA}.

\bibitem[{Park et~al.(2020)Park, Zhang, Jia, Han, Chiu, Li, Wu, and
  Le}]{park2020improved}
Daniel~S. Park, Yu~Zhang, Ye~Jia, Wei Han, Chung{-}Cheng Chiu, Bo~Li, Yonghui
  Wu, and Quoc~V. Le. 2020.
\newblock \href {https://doi.org/10.21437/Interspeech.2020-1470} {Improved
  noisy student training for automatic speech recognition}.
\newblock In \emph{Interspeech 2020, 21st Annual Conference of the
  International Speech Communication Association, Virtual Event, Shanghai,
  China, 25-29 October 2020}, pages 2817--2821. {ISCA}.

\bibitem[{Parthasarathi and Strom(2019)}]{parthasarathi2019lessons}
Sree Hari~Krishnan Parthasarathi and Nikko Strom. 2019.
\newblock \href {https://doi.org/10.1109/ICASSP.2019.8683690} {Lessons from
  building acoustic models with a million hours of speech}.
\newblock In \emph{{IEEE} International Conference on Acoustics, Speech and
  Signal Processing, {ICASSP} 2019, Brighton, United Kingdom, May 12-17, 2019},
  pages 6670--6674. {IEEE}.

\bibitem[{Pasad et~al.(2021)Pasad, Chou, and Livescu}]{pasad2021layer}
Ankita Pasad, Ju-Chieh Chou, and Karen Livescu. 2021.
\newblock Layer-wise analysis of a self-supervised speech representation model.
\newblock In \emph{ASRU}.

\bibitem[{Pradhan et~al.(2013)Pradhan, Moschitti, Xue, Ng, Bj{\"o}rkelund,
  Uryupina, Zhang, and Zhong}]{pradhan2013towards}
Sameer Pradhan, Alessandro Moschitti, Nianwen Xue, Hwee~Tou Ng, Anders
  Bj{\"o}rkelund, Olga Uryupina, Yuchen Zhang, and Zhi Zhong. 2013.
\newblock \href {https://aclanthology.org/W13-3516} {Towards robust linguistic
  analysis using {O}nto{N}otes}.
\newblock In \emph{Proceedings of the Seventeenth Conference on Computational
  Natural Language Learning}, pages 143--152, Sofia, Bulgaria. Association for
  Computational Linguistics.

\bibitem[{Ratinov and Roth(2009)}]{ratinov2009design}
Lev Ratinov and Dan Roth. 2009.
\newblock \href {https://aclanthology.org/W09-1119} {Design challenges and
  misconceptions in named entity recognition}.
\newblock In \emph{Proceedings of the Thirteenth Conference on Computational
  Natural Language Learning ({C}o{NLL}-2009)}, pages 147--155, Boulder,
  Colorado. Association for Computational Linguistics.

\bibitem[{Raymond(2013)}]{raymond2013robust}
Christian Raymond. 2013.
\newblock Robust tree-structured named entities recognition from speech.
\newblock In \emph{ICASSP}.

\bibitem[{Riloff(1996)}]{riloff1996automatically}
Ellen Riloff. 1996.
\newblock Automatically generating extraction patterns from untagged text.
\newblock In \emph{AAAI}.

\bibitem[{Schneider et~al.(2019)Schneider, Baevski, Collobert, and
  Auli}]{schneider2019wav2vec}
Steffen Schneider, Alexei Baevski, Ronan Collobert, and Michael Auli. 2019.
\newblock \href {https://doi.org/10.21437/Interspeech.2019-1873} {wav2vec:
  Unsupervised pre-training for speech recognition}.
\newblock In \emph{Interspeech 2019, 20th Annual Conference of the
  International Speech Communication Association, Graz, Austria, 15-19
  September 2019}, pages 3465--3469. {ISCA}.

\bibitem[{Scudder(1965)}]{scudder1965probability}
Henry Scudder. 1965.
\newblock Probability of error of some adaptive pattern-recognition machines.
\newblock \emph{IEEE Transactions on Information Theory}.

\bibitem[{Shon et~al.(2022)Shon, Pasad, Wu, Brusco, Artzi, Livescu, and
  Han}]{slue}
Suwon Shon, Ankita Pasad, Felix Wu, Pablo Brusco, Yoav Artzi, Karen Livescu,
  and Kyu~J Han. 2022.
\newblock {SLUE}: {N}ew benchmark tasks for spoken language understanding
  evaluation on natural speech.
\newblock In \emph{ICASSP}.

\bibitem[{Sudoh et~al.(2006)Sudoh, Tsukada, and
  Isozaki}]{sudoh2006incorporating}
Katsuhito Sudoh, Hajime Tsukada, and Hideki Isozaki. 2006.
\newblock \href {https://doi.org/10.3115/1220175.1220253} {Incorporating speech
  recognition confidence into discriminative named entity recognition of speech
  data}.
\newblock In \emph{Proceedings of the 21st International Conference on
  Computational Linguistics and 44th Annual Meeting of the Association for
  Computational Linguistics}, pages 617--624, Sydney, Australia. Association
  for Computational Linguistics.

\bibitem[{Wang et~al.(2021{\natexlab{a}})Wang, Riviere, Lee, Wu, Talnikar,
  Haziza, Williamson, Pino, and Dupoux}]{wang2021voxpopuli}
Changhan Wang, Morgane Riviere, Ann Lee, Anne Wu, Chaitanya Talnikar, Daniel
  Haziza, Mary Williamson, Juan Pino, and Emmanuel Dupoux. 2021{\natexlab{a}}.
\newblock \href {https://doi.org/10.18653/v1/2021.acl-long.80} {{V}ox{P}opuli:
  A large-scale multilingual speech corpus for representation learning,
  semi-supervised learning and interpretation}.
\newblock In \emph{Proceedings of the 59th Annual Meeting of the Association
  for Computational Linguistics and the 11th International Joint Conference on
  Natural Language Processing (Volume 1: Long Papers)}, pages 993--1003,
  Online. Association for Computational Linguistics.

\bibitem[{Wang et~al.(2021{\natexlab{b}})Wang, Jiang, Bach, Wang, Huang, Huang,
  and Tu}]{wang2020automated}
Xinyu Wang, Yong Jiang, Nguyen Bach, Tao Wang, Zhongqiang Huang, Fei Huang, and
  Kewei Tu. 2021{\natexlab{b}}.
\newblock \href {https://doi.org/10.18653/v1/2021.acl-long.206} {Automated
  concatenation of embeddings for structured prediction}.
\newblock In \emph{Proceedings of the 59th Annual Meeting of the Association
  for Computational Linguistics and the 11th International Joint Conference on
  Natural Language Processing (Volume 1: Long Papers)}, pages 2643--2660,
  Online. Association for Computational Linguistics.

\bibitem[{Wolf et~al.(2020)Wolf, Debut, Sanh, Chaumond, Delangue, Moi, Cistac,
  Rault, Louf, Funtowicz et~al.}]{wolf2019huggingface}
Thomas Wolf, Lysandre Debut, Victor Sanh, Julien Chaumond, Clement Delangue,
  Anthony Moi, Pierric Cistac, Tim Rault, R{\'e}mi Louf, Morgan Funtowicz,
  et~al. 2020.
\newblock Huggingface's transformers: {S}tate-of-the-art natural language
  processing.
\newblock In \emph{EMNLP}.

\bibitem[{Xu et~al.(2021)Xu, Baevski, Likhomanenko, Tomasello, Conneau,
  Collobert, Synnaeve, and Auli}]{xu2021self}
Qiantong Xu, Alexei Baevski, Tatiana Likhomanenko, Paden Tomasello, Alexis
  Conneau, Ronan Collobert, Gabriel Synnaeve, and Michael Auli. 2021.
\newblock Self-training and pre-training are complementary for speech
  recognition.
\newblock In \emph{ICASSP}.

\bibitem[{Xu et~al.(2020)Xu, Likhomanenko, Kahn, Hannun, Synnaeve, and
  Collobert}]{xu2020iterative}
Qiantong Xu, Tatiana Likhomanenko, Jacob Kahn, Awni Hannun, Gabriel Synnaeve,
  and Ronan Collobert. 2020.
\newblock \href {https://doi.org/10.21437/Interspeech.2020-1800} {Iterative
  pseudo-labeling for speech recognition}.
\newblock In \emph{Interspeech 2020, 21st Annual Conference of the
  International Speech Communication Association, Virtual Event, Shanghai,
  China, 25-29 October 2020}, pages 1006--1010. {ISCA}.

\bibitem[{Yadav et~al.(2020)Yadav, Ghosh, Yu, and Shah}]{yadav2020end}
Hemant Yadav, Sreyan Ghosh, Yi~Yu, and Rajiv~Ratn Shah. 2020.
\newblock \href {https://doi.org/10.21437/Interspeech.2020-2482} {End-to-end
  named entity recognition from english speech}.
\newblock In \emph{Interspeech 2020, 21st Annual Conference of the
  International Speech Communication Association, Virtual Event, Shanghai,
  China, 25-29 October 2020}, pages 4268--4272. {ISCA}.

\bibitem[{Yadav and Bethard(2018)}]{yadav2019survey}
Vikas Yadav and Steven Bethard. 2018.
\newblock \href {https://aclanthology.org/C18-1182} {A survey on recent
  advances in named entity recognition from deep learning models}.
\newblock In \emph{Proceedings of the 27th International Conference on
  Computational Linguistics}, pages 2145--2158, Santa Fe, New Mexico, USA.
  Association for Computational Linguistics.

\bibitem[{Yang et~al.(2021)Yang, Chi, Chuang, Lai, Lakhotia, Lin, Liu, Shi,
  Chang, Lin, Huang, Tseng, Lee, Liu, Huang, Dong, Li, Watanabe, Mohamed, and
  Lee}]{yang2021superb}
Shu{-}Wen Yang, Po{-}Han Chi, Yung{-}Sung Chuang, Cheng{-}I~Jeff Lai, Kushal
  Lakhotia, Yist~Y. Lin, Andy~T. Liu, Jiatong Shi, Xuankai Chang, Guan{-}Ting
  Lin, Tzu{-}Hsien Huang, Wei{-}Cheng Tseng, Ko{-}tik Lee, Da{-}Rong Liu, Zili
  Huang, Shuyan Dong, Shang{-}Wen Li, Shinji Watanabe, Abdelrahman Mohamed, and
  Hung{-}yi Lee. 2021.
\newblock \href {https://doi.org/10.21437/Interspeech.2021-1775} {{SUPERB:}
  speech processing universal performance benchmark}.
\newblock In \emph{Interspeech 2021, 22nd Annual Conference of the
  International Speech Communication Association, Brno, Czechia, 30 August - 3
  September 2021}, pages 1194--1198. {ISCA}.

\bibitem[{Yarowsky(1995)}]{yarowsky1995unsupervised}
David Yarowsky. 1995.
\newblock \href {https://doi.org/10.3115/981658.981684} {Unsupervised word
  sense disambiguation rivaling supervised methods}.
\newblock In \emph{33rd Annual Meeting of the Association for Computational
  Linguistics}, pages 189--196, Cambridge, Massachusetts, USA. Association for
  Computational Linguistics.

\end{thebibliography}
\bibliographystyle{acl_natbib}

\vfill\pagebreak
\appendix
\section{Appendix}
\label{sec:appendix}
\subsection{Results on the test set}
\label{app:test-results}

\begin{figure}[h]
    \centering
    \includegraphics[width=8cm, trim=0 65 0 85, clip]{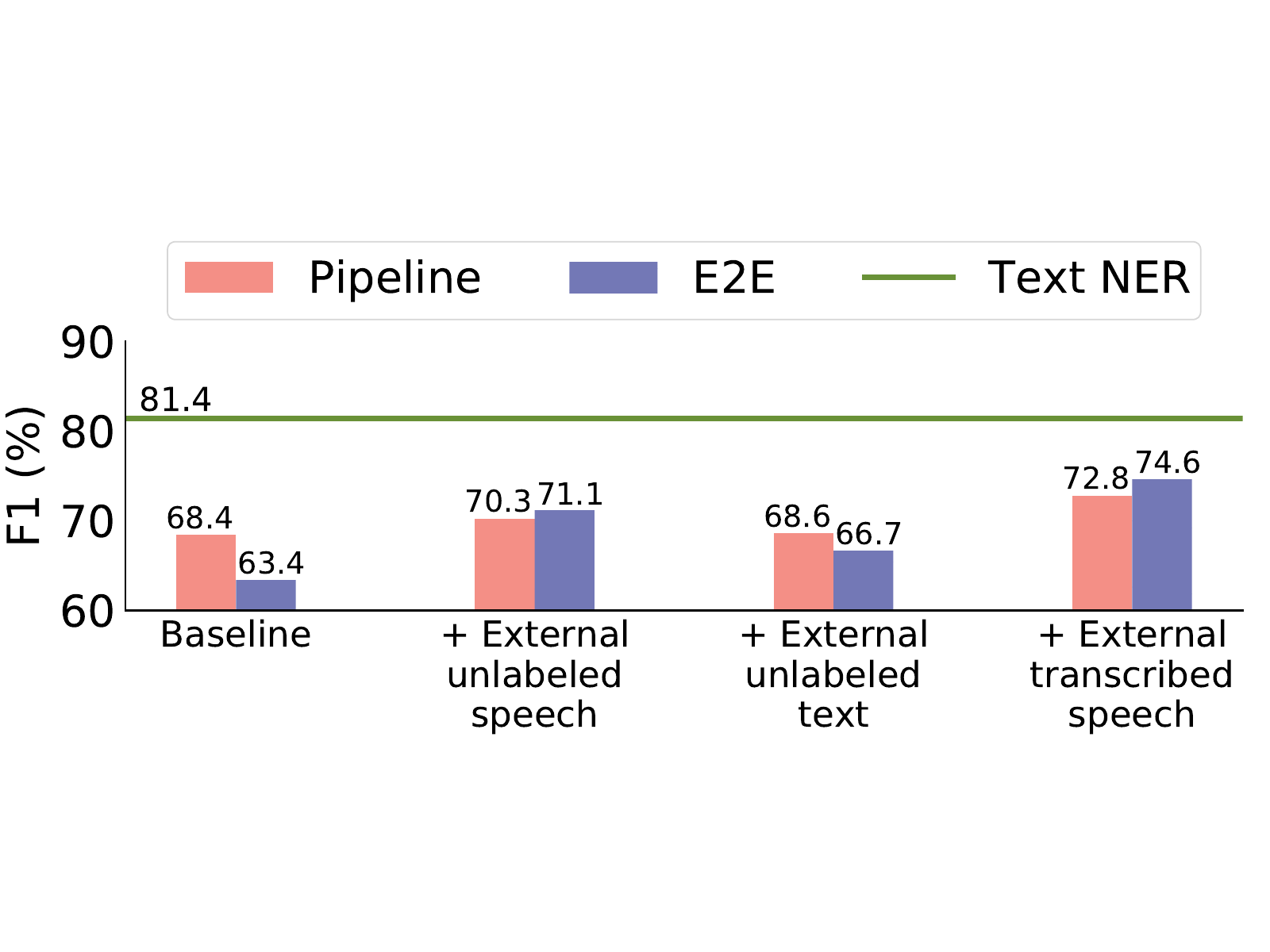}
    \caption{\it Spoken NER test set results with 100 hours of external data of different types. The ``Baseline'' and ``Text NER'' numbers are from~\citet{slue}.}
    \label{fig:summary-test}
\end{figure}

 We obtain test set results for our best-performing models, by submitting model outputs following the SLUE instructions.\footnote{\href{https://asappresearch.github.io/slue-toolkit/how-to-submit.html}{https://asappresearch.github.io/slue-toolkit/how-to-submit.html}}. These results are presented in Fig.~\ref{fig:summary-test}. We observe similar trends as on the dev set (see Fig.~\ref{fig:summary}).

\begin{figure}[h]
% {<left> <lower> <right> <upper>}
\begin{minipage}[b]{1.0\linewidth}
\small
% \vspace{-0.5cm}
 \centering
 \centerline{\includegraphics[width=8cm, trim=0 75 0 100, clip]{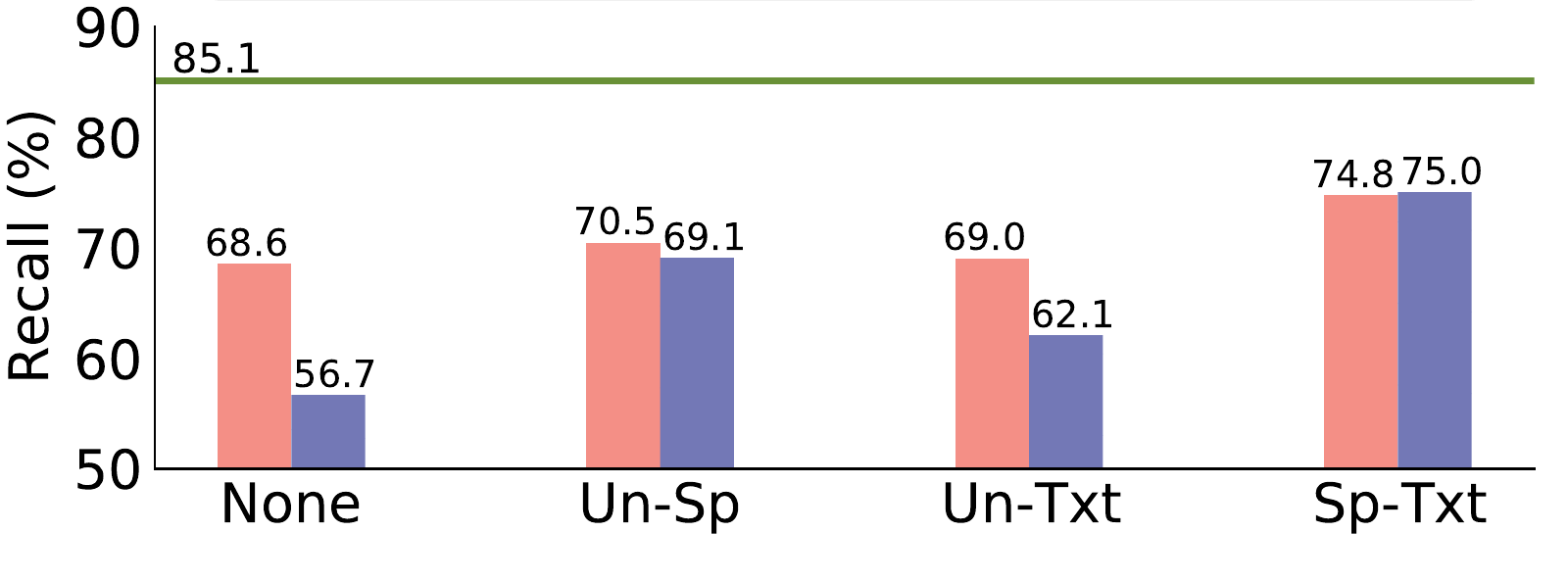}}
\end{minipage}
\begin{minipage}[b]{1.0\linewidth}

\vspace{-0.05cm}
\footnotesize
 \centering
  \centerline{\includegraphics[width=8cm, trim=0 65 0 75, clip]{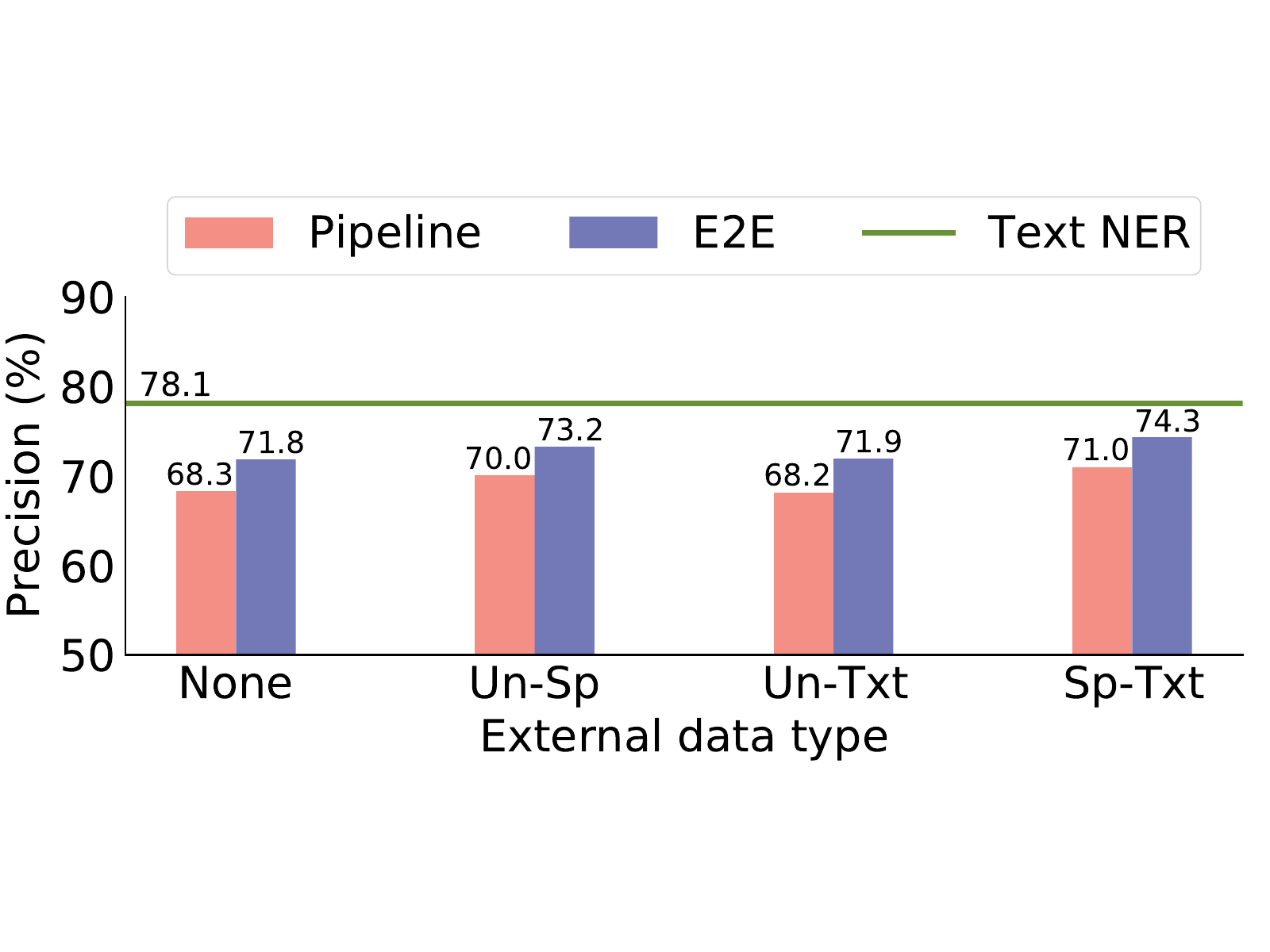}}
\end{minipage}

\vspace{-0.3cm}
\caption{{\it Recall and precision on the test set for the best performing models using 100 hours of external data.}}
  \label{fig:prec-recall-test}
\vspace{-0.3cm}
\end{figure}

We can see from the precision and recall scores in  Fig.~\ref{fig:prec-recall-test}, that our analytical conclusions about the pipeline model performing poorly due to false positives are consistent across these two splits.

\subsection{Error categories}
\label{app:error-cat}
\begin{figure*}[h]
% {<left> <lower> <right> <upper>}
\begin{minipage}[b]{1.0\linewidth}
\small
% \vspace{-0.5cm}
 \centering
 \centerline{\includegraphics[width=\linewidth]{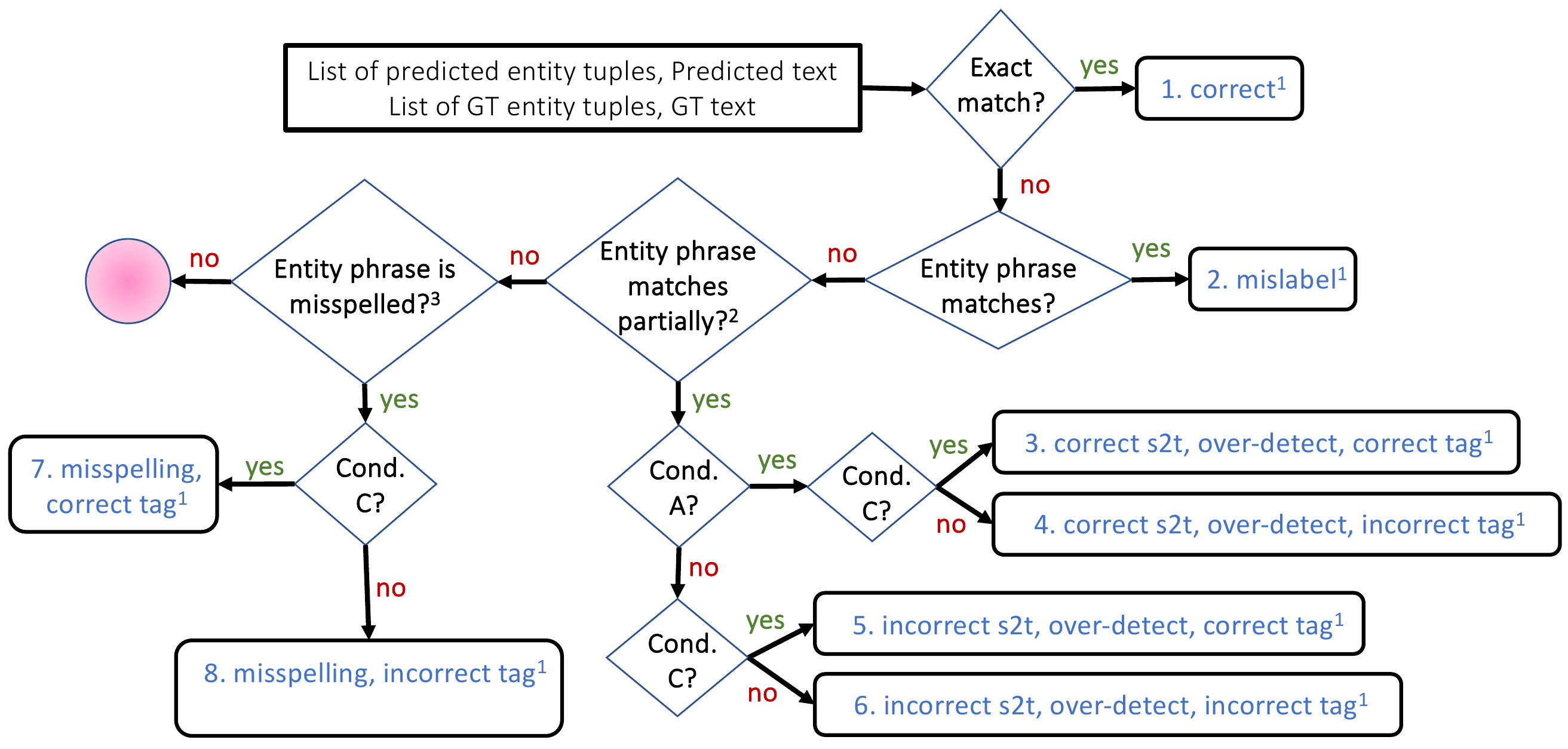}}
\end{minipage}
\begin{minipage}[b]{1.0\linewidth}

\vspace{1cm}
\footnotesize
 \centering
  \centerline{\includegraphics[width=\linewidth]{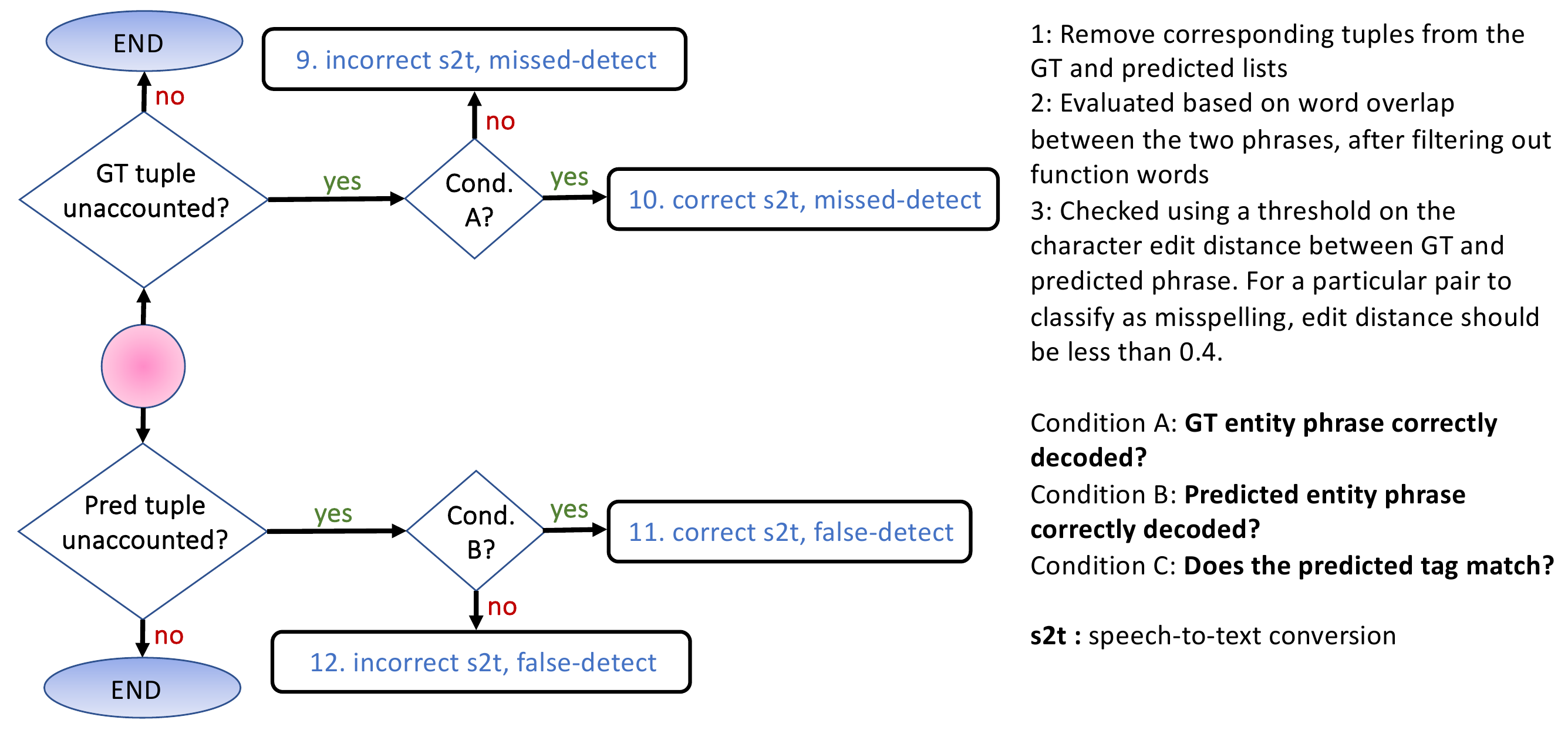}}
\end{minipage}

\vspace{-0.3cm}
\caption{{\it Illustration of algorithm for obtaining error category types for each (entity phrase, entity tag) tuple in ground-truth and predicted outputs.}}
  \label{fig:flowchart}
\vspace{-0.3cm}
\end{figure*}

Fig.~\ref{fig:flowchart} illustrates via a flowchart our process of assigning the tuples in ground-truth and predicted outputs into different error categories.
% \usepackage{multirow}
\begin{table*}[]
\normalsize
\centering
\begin{tabular}{lcc}
\multicolumn{1}{c|}{\multirow{2}{*}{\textbf{Error category}}}                                  & \multicolumn{2}{c}{\textbf{Outputs from E2E model}}                                                                                                                                                                                                                                                                                                                                                                                                                                                                                                                                                                                      \\ \cline{2-3}
\multicolumn{1}{c|}{}                                                                          & \textbf{GT}                                                                                                                                                                                                                                                                                                                           & \textbf{Predicted}                                                                                                                                                                                                                                                                               \\ \midrule
\multicolumn{1}{l|}{\begin{tabular}[c]{@{}l@{}}Correct ASR,\\ over detection\end{tabular}}    & \multicolumn{1}{c|}{\begin{tabular}[c]{@{}c@{}}and this means that you look\\  and tell us honestly what does it \\ mean if you start @ three years {]} \\ later\\ \\ {[}(`WHEN', `three years'){]}\end{tabular}}                                                                                                                     & \begin{tabular}[c]{@{}c@{}}this means that you look and \\ tell us honestly what does it mean if\\ you start @ three years later {]}\\ \\ {[}(`WHEN', `three years later'){]}\end{tabular}                                                                                                       \\ \midrule
\multicolumn{1}{l|}{\begin{tabular}[c]{@{}l@{}}Correct ASR, \\ missed detection\end{tabular}} & \multicolumn{1}{c|}{\begin{tabular}[c]{@{}c@{}}the situation in the \% drc {]} is indeed \\ terrible and it has been this way for \\ quite a while and i am deeply \\ concerned about the handling of the \\ current issue with regard to the \% kasai {]} \\ province \\ \\ {[}(`PLACE', `drc'), (`PLACE', `kasai'){]}\end{tabular}} & \begin{tabular}[c]{@{}c@{}}the situation in the drc is indeed \\ terrible and it has been this way for \\ for quite a while and i am deeply \\ concerned about the handling of\\ the current issue with regard to \\ the a province \\ \\ {[}{]}\end{tabular}                                    \\ \midrule
\multicolumn{1}{l|}{\begin{tabular}[c]{@{}l@{}}Correct ASR,\\ false detection\end{tabular}}   & \multicolumn{1}{c|}{\begin{tabular}[c]{@{}c@{}}and yet @ one month {]} after we adopted \\ our compromise the council did not \\ put it on the agenda did not even present\\  it i used this time to talk to the \\ member states and the presidencies\\ \\ {[}(`WHEN', `one month'){]}\end{tabular}}                                 & \begin{tabular}[c]{@{}c@{}}still @ one month {]} after we voted \\ a compromise the ` council {]} did \\ not put it on the agenda did \\ not even present i use this time \\ to talk with the member states and \\ the presidency \\ \\ {[}(`WHEN', `one month'), \\ (`ORG', `council'){]}\end{tabular} \\ \midrule
\multicolumn{1}{l|}{\begin{tabular}[c]{@{}l@{}}Incorrect ASR,\\ false detection\end{tabular}} & \multicolumn{1}{c|}{\begin{tabular}[c]{@{}c@{}}it has nothing to do with religion \\ but it has all to do with patriarchy\\ \\ {[}{]}\end{tabular}}                                                                                                                                                                                   & \begin{tabular}[c]{@{}c@{}}it has nothing to do with religion \\ but it has all to do with \% turkey {]}\\ \\ {[}(`PLACE', `turkey'){]}\end{tabular}                                                                                                                                             \\ \midrule
\end{tabular}
\caption{Qualitative examples for different error categories from the output of the E2E model using 100 hours of unlabeled speech (\it Distill-Pipeline).}
\label{tab:qualitative-examples}
\end{table*}
Tab.~\ref{tab:qualitative-examples} presents examples for the four categories discussed in Sec~\ref{sec:analysis}. These are examples from the dev set, using the {\it Distill-Pipeline} E2E model trained on 100 hours of data.

\end{document}